\documentclass[final,1p,times,authoryear]{elsarticle}
\usepackage[utf8]{inputenc}

\usepackage{float}
\usepackage{amssymb}
\usepackage{amsmath}
\usepackage{booktabs}
\usepackage{tabularx}
\usepackage[colorlinks]{hyperref}
\usepackage{natbib}
\bibpunct{[}{]}{,}{n}{}{;}
\usepackage{longtable}
\usepackage{makecell}
\usepackage{caption}
\usepackage{setspace}
\biboptions{sort&compress}

\newcolumntype{Y}{>{\raggedright\arraybackslash}X}
\journal{Journal of Biomedical Informatics}

\begin{document}

\begin{frontmatter}







\title{A Scoping Review of Natural Language Processing in Addressing Medically Inaccurate Information: Errors, Misinformation, and Hallucination}

    
\author[1]{Zhaoyi Sun\corref{cor1}}
\ead{zhaoyis@uw.edu}
\author[2]{Wen-Wai Yim}
\author[3]{Özlem Uzuner} 
\author[4]{Fei Xia} 
\author[1]{Meliha Yetisgen}   

\cortext[cor1]{Corresponding author}
    
\affiliation[1]{
    organization={Biomedical Informatics and Medical Education},
    addressline={University of Washington}, 
    city={Seattle},      
    state={WA},
    postcode={98195},
    country={USA}}

\affiliation[2]{
    organization={Health AI},
    addressline={Microsoft},
    city={Redmond}, 
    state={WA},
    postcode={98052},
    country={USA}}

\affiliation[3]{
    organization={Department of Information Sciences and Technology},
    addressline={George Mason University},
    city={Fairfax},  
    state={VA},
    postcode={22030},
    country={USA}}

\affiliation[4]{
    organization={Department of Linguistics},
    addressline={University of Washington}, 
    city={Seattle},  
    state={WA},
    postcode={98195},
    country={USA}}

\begin{abstract}
\textbf{Objective:} This review aims to explore the potential and challenges of using Natural Language Processing (NLP) to detect, correct, and mitigate medically inaccurate information, including errors, misinformation, and hallucination. By unifying these concepts, the review emphasizes their shared methodological foundations and their distinct implications for healthcare. Our goal is to advance patient safety, improve public health communication, and support the development of more reliable and transparent NLP applications in healthcare.

\noindent
\textbf{Methods:} A scoping review was conducted following PRISMA-ScR guidelines, analyzing studies from 2020 to 2024 across five databases. Studies were selected based on their use of NLP to address medically inaccurate information and were categorized by topic, tasks, document types, datasets, models, and evaluation metrics.

\noindent
\textbf{Results:} NLP has shown potential in addressing medically inaccurate information on the following tasks: (1) error detection (2) error correction (3) misinformation detection (4) misinformation correction (5) hallucination detection (6) hallucination mitigation. However, challenges remain with data privacy, context dependency, and evaluation standards.

\noindent
\textbf{Conclusion:} This review highlights the advancements in applying NLP to tackle medically inaccurate information while underscoring the need to address persistent challenges. Future efforts should focus on developing real-world datasets, refining contextual methods, and improving hallucination management to ensure reliable and transparent healthcare applications.

\end{abstract}




\end{frontmatter}


\section{Introduction}

Medically inaccurate information refers to incorrect text data or communication related to health and medicine. It can be categorized as errors, misinformation and hallucination based on the source of the information and its disseminability. Errors represent inaccurate information in clinical texts, such as electronic health records (EHRs), clinical notes, and patient reports. These inaccuracies typically remain contained within healthcare systems and do not spread widely. However, their impact can be profound, leading to medication mistakes, misdiagnoses, inappropriate treatments, and adverse patient outcomes~\cite{Assiri2018-da, Legat2018-wl, Al-Meslamani2023-lf}. In contrast, misinformation is a category of inaccurate information with the potential to disseminate widely, leading to harmful health behaviors and eroding trust in healthcare providers~\cite{Joseph2022-ad}. For example, during the COVID-19 pandemic, the proportion of social media posts containing COVID-19-related misinformation was as high as 28.8$\%$~\cite{Gabarron2021-ph}. Misinformation has led to vaccine hesitancy and denial of the severity of COVID-19 infection~\cite{Joseph2022-ad, Clemente-Suarez2022-uy}. 

In this article, misinformation includes both unintentional inaccuracies and intentionally misleading content. This definition takes into consideration the ongoing debate about the definition of misinformation. Some studies use this term specifically to refer to unintentional inaccuracies~\cite{Armitage2021-ib,Soe2021-uo,Pennycook2020-rh,Baines2020-zw}. Unintentional inaccuracies often arise from misunderstandings or misinterpretations and are shared without any intent to deceive. Intentially misleading content is often referred to as disinformation. Disinformation is usually driven by personal, political, or financial motives~\cite{wardle2018information}.  In the absence of information about the intent of the author, misinformation and disinformation are difficult to distinguish from each other.  Detection of author intent is outside the scope of this article; therefore, in this paper the term ``misinformation'' refers to both unintentional and intentional inaccuracies. 

\begin{figure}[htbp]
\includegraphics[width=\textwidth]{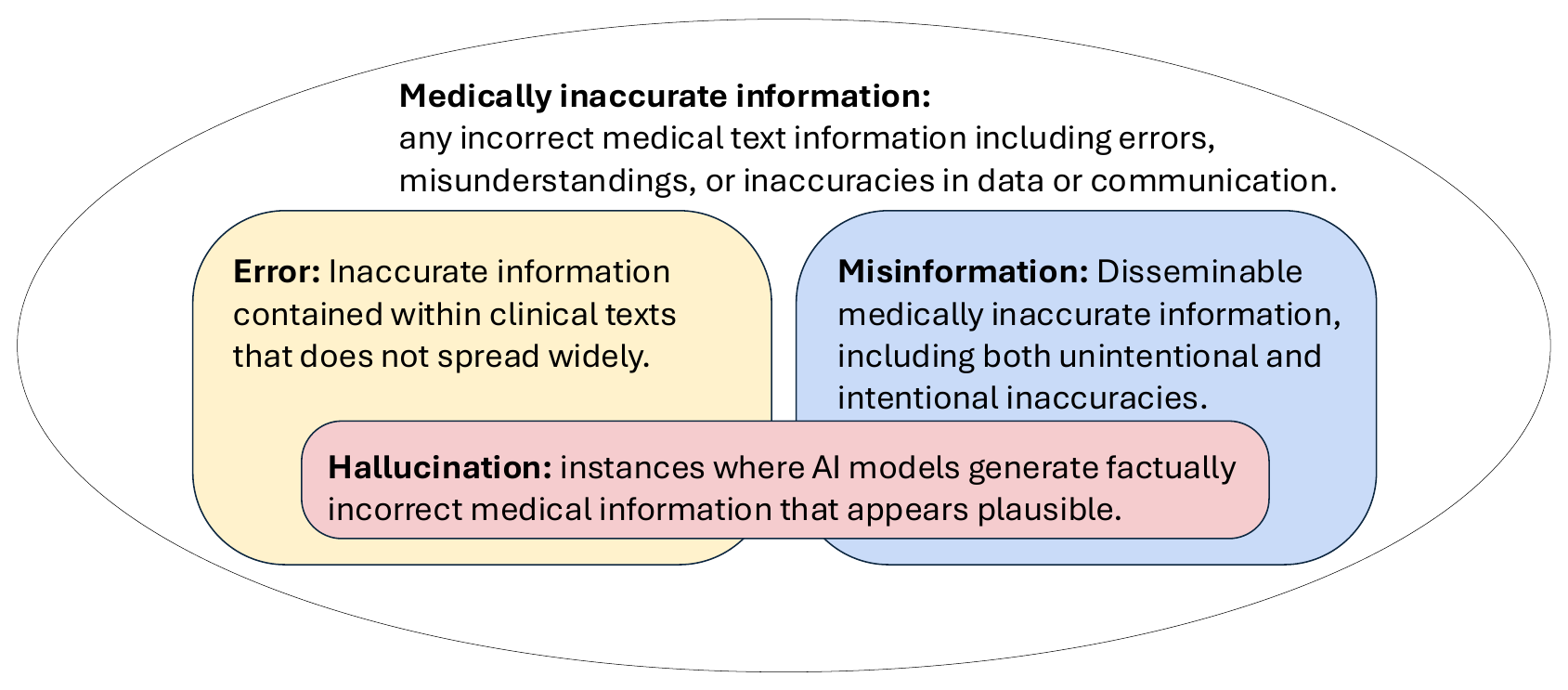}
\caption{Venn diagram of medically inaccurate information, error, misinformation, and hallucination}
\label{fig:venn_diagram_medical_wrong_information}
\end{figure}

With the rise of Artificial Intelligence (AI) and Large Language Models (LLMs) in the medical domain, a new source of medically inaccurate information has emerged: hallucination. Hallucination
occurs when AI generates incorrect information that appears plausible and contextually fitting~\cite{Zhang2023-qk}. Figure \ref{fig:venn_diagram_medical_wrong_information} presents medically inaccurate information, errors, misinformation and hallucination in a Venn diagram. Hallucination can overlap with both errors and misinformation. For example, when LLMs are used to generate clinical texts, hallucination can add inaccurate details to clinical notes, and may misguide clinical decisions~\cite{Goodman2024-cr}. Similarly, LLMs may affect diagnostic accuracy by misinterpreting lab results due to limitations in contextual understanding~\cite{Ullah2024-ny}. Additionally, when LLMs are employed for patient education or public health messaging, hallucination can unintentionally disseminate misinformation. For instance, an AI-generated patient education material may inaccurately imply universal safety without considering individual health conditions~\cite{Aydin2024-jk}. AI-generated health messages for public awareness may also occasionally include outdated or overly generalized information~\cite{Lim2023-oo}. Although there is no evidence to suggest that LLMs can intentionally generate disinformation, they remain susceptible to manipulation. Altering as little as 1.1$\%$ of their weights has been shown to introduce and propagate inaccurate biomedical facts~\cite{Han2024-ia}. Furthermore, the generation speed of LLMs is remarkable, with the ability to produce 102 misleading blog posts with fabricated references in just 65 minutes~\cite{Menz2024-hl}. However, current legal and regulatory measures are inadequate to effectively address these vulnerabilities effectively~\cite{Menz2024-dx}. Therefore, it is essential to recognize and mitigate AI-induced hallucination to prevent dissemination of medically inaccurate information and to ensure the reliability of NLP in healthcare.

Advancements in Natural Language Processing (NLP), especially LLMs, are revolutionizing the way we detect and correct medically inaccurate information. Transformer-based models like Bidirectional Encoder Representations from Transformers (BERT)~\cite{Devlin2018-vb} and their medical domain adaptations, such as BioBERT~\cite{Lee2020-xg} and ClinicalBERT~\cite{Huang2019-ha}, have been utilized for error detection and classification in clinical texts~\cite{Lee2022-hk}. NLP techniques also played a critical role in misinformation detection by monitoring social media and online forums for health-related inaccuracies. During the COVID-19 pandemic, models were trained to detect COVID-19-related misinformation from tweets by classifying content based on veracity~\cite{Alam2021-cb}. Additionally, tasks such as fact-checking and claim verification leveraged NLP combined with knowledge graphs to verify claims against trusted medical databases, including PubMed and guidelines from the Centers for Disease Control and Prevention (CDC)~\cite{Sarrouti2021-cq, Vladika2024-wl}. Although encoder-only models like BERT can also be referred to as LLMs~\cite{sindhu2024evolution,shool2025systematic}, we followed the more common usage that restricts the term ``LLM" for generative models. LLMs such as GPT-4~\cite{OpenAI2022-xn} exhibited unique strengths in error correction by generating contextually rich, human-like explanations and reasoning through interactive dialogues~\cite{Gundabathula2024-vl}. These capabilities, supported by advancements in natural language understanding (NLU), enabled LLMs to integrate domain-specific knowledge and address complex inaccuracies with greater flexibility~\cite{Toma2024-io,Khattab2023-vl}. 

However, as LLMs evolve, their ability to generate fluent and contextually appropriate text introduces a paradox: models designed to mitigate inaccuracies may inadvertently create new ones. To address these challenges, researchers have developed diverse methods for hallucination detection, evaluation, and mitigation. Standardized evaluation frameworks are emerging to systematically benchmark a model’s factual correctness~\cite{Tam2024-cq}. Additionally, techniques such as retrieval-augmented generation (RAG) and chain-of-thought (CoT) prompting were utilized to improve the accuracy of generated text~\cite{Lievin2024-ae,Alkhalaf2024-nm,Gilbert2024-uv,Tonmoy2024-lj}. Human-in-the-loop systems further enhanced reliability by incorporating expert review of AI-generated materials~\cite{Ahmad2023-zs}. These efforts marked significant progress toward building trustworthy LLMs, though continuous refinement and collaboration remain essential to addressing the complexities of AI-driven text generation.

Our review is inspired by several prior review articles. Schlicht et al.~\cite{Schlicht2023-ke} conducted a comprehensive analysis of misinformation detection in healthcare, categorizing techniques and datasets across various misinformation themes, particularly in relation to COVID-19. However, they omitted the important aspects of errors and hallucination in clinical contexts. Suarez-Lledo et al.~\cite{Suarez-Lledo2021-pg} examined prevalence of health misinformation about vaccines and non-communicable diseases on social media, but their focus was primarily on the dissemination dynamics rather than on the technical methods for detection and correction. Su et al.~\cite{Su2020-ce} reviewed misinformation detection techniques from an NLP perspective, discussing motivations, methods, and metrics in the field. However, their review predates the emergence of LLMs and does not address the significant advancements introduced by these models.  Chen et al.~\cite{Chen2024-ox} examined the dual role of LLMs in both misinformation detection and hallucination generation. Yet, their analysis primarily centered on general-domain LLMs, without considering traditional methods for medical error and misinformation detection.

To our knowledge, our review is the first study focusing on NLP in addressing the full spectrum of medically inaccurate information - errors, misinformation, and hallucination. By unifying these concepts, we highlight their shared methodological foundations in NLP while emphasizing their distinct implications for healthcare. The motivation of this review is to advance the detection, correction, and mitigation of medically inaccurate information and improve accuracy, reliability, and safety in healthcare information systems. Our target readers include computer scientists, healthcare professionals, medical journalists, and policymakers. Specifically, computer scientists can leverage our findings to develop more robust NLP algorithms tailored to medical data with enhanced transparency and explainability; healthcare professionals can apply these techniques to clinical practice to improve patient safety by uncovering potential errors in clinical notes and identifying misinformation in patient education materials; medical journalists can improve public health communication by adopting fact-checked, accurate information dissemination practices, and carefully examining content with political or financial motivations, as such information is more likely to carry disinformation~\cite{wardle2017information}; policymakers can use insights from NLP advancements to guide regulations that support reliable health information systems. Joint efforts among technical experts, healthcare professionals, communicators, and policymakers should be made to combat medically inaccurate information.

The primary review question in this study is: What is the current state of NLP in addressing medically inaccurate information? To answer this question, we focus on several sub-questions as follows: What datasets were used in relevant studies? What topics were the studies addressing? What specific NLP tasks were explored? What source document types were used? What NLP models or methods were applied? What metrics were utilized to evaluate performance? Was expert evaluation involved in assessing the model performance? These sub-questions frame our scoping review of current methodologies and resources, with the goal of identifying strengths, gaps, and future directions for enhancing NLP applications in the detection, correction, and mitigation of medical errors, misinformation, and hallucination.

We structured our review as follows: Section \ref{sec:methods} describes the article selection process, following the protocol for identifying and including relevant studies. Section \ref{sec:results} presents the results of article selection for this review, with an analysis of datasets. Section \ref{sec:errors}-\ref{sec:hallucination} explains the collected papers by NLP tasks, including detection, correction and mitigation for errors, misinformation, and hallucination. Section \ref{sec:discussion} examines the limitations, gaps, and future directions in the methods that aim to address each type of inaccurate information. Section \ref{sec:limitation} discusses the limitations of this scoping review. Finally, Section \ref{sec:conclusion} offers a concise conclusion.

\begin{center}
    \begin{tabular}{p{4cm}p{9cm}}
        \toprule
        \textbf{Problems} & Medically inaccurate information (errors, misinformation, hallucination) impacts healthcare reliability and safety. Errors in clinical texts can result in incorrect diagnoses and treatments. Misinformation spreads broadly, causing harmful health behaviors and reducing trust in healthcare. Hallucination from LLMs creates plausible but inaccurate information, complicating decisions and communication. \\\\
        \textbf{What is Already Known} & NLP has shown promise in detecting and correcting errors and misinformation. The advancement of LLMs has improved performance but also introduced new risks: hallucination. \\\\
        \textbf{What this Paper Adds} & This paper unifies the concept of medically inaccurate information, highlighting shared methodological foundations while emphasizing distinctions in datasets, tasks, document types, models, evaluation metrics, and expert involvement. It identifies strengths, gaps, and future directions, providing insights to improve NLP methods for detecting, correcting, and mitigating errors, misinformation, and hallucination in healthcare. \\\\
        \textbf{Who Would Benefit from the New Knowledge} & Computer scientists designing robust NLP models; healthcare professionals aiming to identify errors in clinical texts; medical journalists promoting accurate health messaging; policymakers guiding AI regulations. \\
        \bottomrule
    \end{tabular}
\end{center}

\section{Methods}
\label{sec:methods}

Our scoping review follows the Preferred Reporting Items for Systematic Reviews and Meta-Analyses Extension for Scoping Reviews (PRISMA-ScR) guidelines~\cite{Tricco2018-zk}.

\subsection{Eligibility Criteria}

This scoping review included English-language studies published between January 2020 and November 2024, as the COVID-19 pandemic significantly accelerated the publication of research on medically inaccurate information. The focus was on NLP techniques aimed at detecting, correcting, or mitigating medically inaccurate information, including errors, misinformation, and hallucination. Eligible publications included peer-reviewed journal articles, conference papers, and preprints.

\subsection{Information Sources}

The articles were retrieved from multiple academic databases, including PubMed\footnote{\url{https://pubmed.ncbi.nlm.nih.gov/}}, the Institute of Electrical and Electronics Engineers (IEEE) Xplore Digital Library\footnote{\url{https://ieeexplore.ieee.org/Xplore/home.jsp}}, the Association for Computing Machinery (ACM) Digital Library\footnote{\url{https://dl.acm.org/}}, the ACL Anthology\footnote{\url{https://aclanthology.org/}}, and Google Scholar\footnote{\url{https://scholar.google.com/}}. The most recent search was conducted on November 30th, 2024.

\subsection{Search Strategy}

Our search strategy employed three groups of keywords: inaccuracy types (e.g., errors, misinformation, disinformation, hallucination), medical terms (e.g., medical, clinical, healthcare), and technical terms (e.g., natural language processing, large language models, text mining). These groups were combined to conduct searches across five databases, with keywords within the same group linked using `OR', while keywords across different groups were combined using `AND'. In addition to the database searches, we conducted a manual search on Google Scholar. This manual search utilized a set of highly-cited papers as seed references; we then identified papers that cited these seed papers for further screening. Table \ref{tab:search_query} provides a detailed overview of the search queries and numbers for each database.

\subsection{Study Selection}
\label{sec:study selection}

The GPT-4o API was employed to assist in the title and abstract screening process. The title and abstract of each paper were input into the system with the following prompt: 

\textit{``Answer with yes or no. Determine whether this paper should be included in a scoping review of natural language processing in the detection, correction, and mitigation of medically inaccurate information, including errors in clinical text, misinformation, disinformation, or hallucination, based on the following title and abstract. Exclude papers if they are non-research articles (e.g., reviews, commentaries, or letters to the editor), outside the medical domain, not in English, unrelated to inaccurate information, or lacking NLP methods. The title and abstract are as follows: $<$title$>$ + $<$abstract$>$."} 

Both GPT-4o and a human reviewer (ZS) independently conducted title and abstract screenings. The Cohen's kappa was 0.70, indicating substantial agreement between GPT-4o and ZS. Besides the exclusion criteria mentioned in the prompts above, papers were also excluded if they were inaccessible or published prior to 2020. If ZS and GPT-4o agreed on a paper, the agreed decision was accepted directly. If ZS and GPT-4o disagreed on a paper, ZS conducted a double-check and decided whether to include it in the next stage. To evaluate the reliability of both agreed and disagreed decisions, a random sample of 40 papers was independently assessed by two additional reviewers (OU and MY), including 10 papers from each of the following agreement/disagreement categories: (1) both included, (2) ZS included but GPT-4o did not, (3) GPT-4o included but ZS did not, and (4) both excluded. All decisions in this sample were consistent with those made by ZS. Papers selected during this screening process then moved to a full-text review by ZS. All included papers and their categorization were discussed with all co-authors before finalizing selections.

\subsection{Data Extraction and Synthesis}

We categorized all articles based on the type of inaccuracies: errors, misinformation, and hallucination. For each category, we summarized the topics addressed (e.g., COVID-19, medication, general medical topics), the NLP tasks involved (e.g., detection, correction, and mitigation), the source document types of information (e.g., Twitter/X posts, health news, clinical text), the datasets used, the NLP methods or models applied, and the metrics used to evaluate model performance. Additionally, we kept a note on whether the evaluation was conducted automatically or involved expert assessment.

\section{Results}
\label{sec:results}

\begin{figure}[H]
\centering
\includegraphics[width=.8\textwidth]{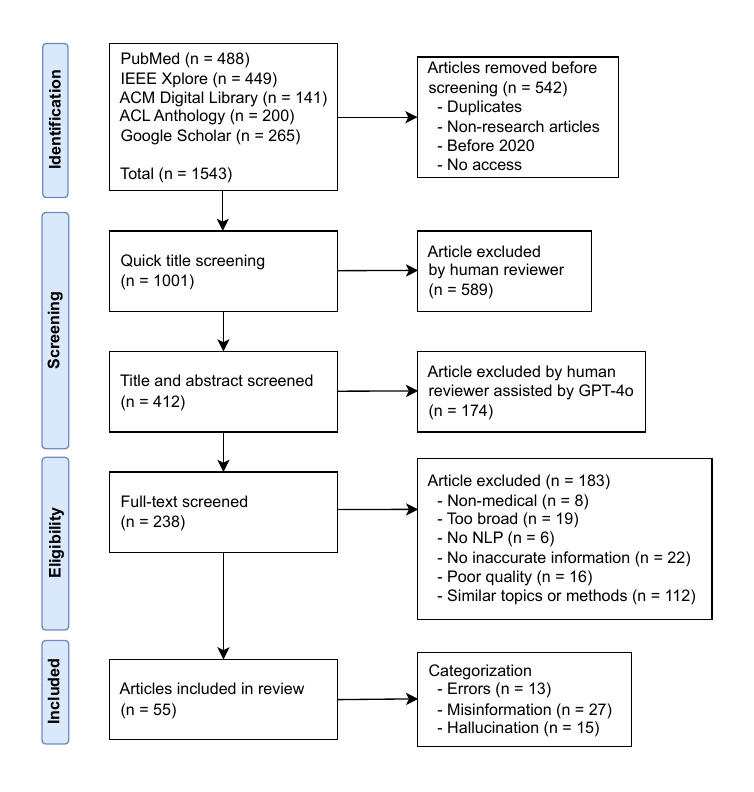}
\caption{Flowchart of article selection following PRISMA guidelines}
\label{fig:article_selection_medical_wrong_information}
\end{figure}

Figure \ref{fig:article_selection_medical_wrong_information} shows the flowchart of article selection, following PRISMA-ScR guidelines. A total of 1,543 articles were initially identified from five databases. Of these, 542 articles were excluded prior to screening for reasons such as duplication, non-research content, publication before 2020, or access issues, leaving 1,001 articles for a quick title screening. Following this initial screening, 589 articles were excluded, and 412 articles proceeded to a detailed title and abstract screening conducted by the human reviewer, assisted by GPT-4o. At this stage, 174 additional articles were excluded based on the criteria outlined in Section \ref{sec:study selection}. Subsequently, 238 articles proceeded to full-text review, during which an additional 183 articles were excluded: 8 were non-medical, 19 had overly broad scopes, 6 lacked a focus on NLP, 22 contained no inaccurate information, and 16 were of poor quality. Articles with ``poor quality" met the initial inclusion criteria but were excluded during full-text review due to issues such as lack of methodological transparency, minimal/vague use of NLP, or absence of NLP evaluation metrics. Additionally, 112 articles were excluded due to overlap in topics or methodologies. For instance, numerous studies on COVID-19-related misinformation detection published between 2020 and 2022 employed similar study designs and models. To avoid redundancy, we reviewed all eligible studies and, when multiple papers shared similar topics, document type, and NLP methods, we randomly selected one to include. Similarly, shared tasks in the field often result in multiple papers addressing the same topic, all meeting our inclusion criteria. In such cases, we included only top 3 ranked submissions. Ultimately, 55 articles were included in our final review, categorized by type of inaccuracy: 13 focused on errors, 27 on misinformation, and 15 on hallucination. 

Table \ref{tab:dataset} summarizes publicly available datasets used or cited in collected articles. The columns contain dataset name, topic, inaccuracy type, source, language, modality, labels and URL. In the following sections, we will provide a detailed exploration of the studies categorized by each type of inaccuracy.

\tiny
\setlength{\tabcolsep}{3pt}
\captionsetup{font=footnotesize}
\begin{longtable}{@{}p{8em}p{5em}p{6em}p{10em}p{5em}p{5em}p{14em}p{14em}@{}}
\caption{Medically inaccurate information datasets (Inaccuracy type: 1 = errors, 2 = misinformation, 3 = hallucination)}
\label{tab:dataset}\\
\toprule
Dataset & Topic & Inaccuracy type & Source & Language & Modality & Labels & URL\\
\midrule
\endfirsthead

\toprule
Dataset & Topic & Inaccuracy type & Source & Language & Modality & Labels & URL\\
\midrule
\endhead

\midrule
\multicolumn{8}{r}{\textit{Continued on next page}}
\endfoot

\bottomrule
\endlastfoot
machine-annotated incident reports of medication errors~\cite{Wong2024-lu} & drug administration & 1                         & incident reports                                                                                              & Japanese                                             & text                              & Intended and Actual, Intended and Not Actual, and Not Intended and Actual                                                                          & \url{https://doi.org/10.6084/m9.figshare.21541650.v3}                            \\\\
Spanish real word error dataset~\cite{Bravo-Candel2021-ko}                         & general medical     & 1                         & clinical case reports with synthetic errors                                                                   & Spanish                                              & text                              & -                                                                                                                                                  & \url{https://pln.inf.um.es/corpora/realworderrors/datasets.rar}                  \\\\
MEDEC~\cite{Abacha2024-se}                                        & general medical     & 1                         & medical question-answering text and clinical notes with injected errors                                       & English                                              & text                              & binary                                                                                                                                             & \url{https://github.com/abachaa/MEDIQA-CORR-2024}                                \\\\
COVID-Lies~\cite{Hossain2020-bc}                                        & COVID-19     & 2                         & Twitter/X (tweets)                                       & English                                              & text                              & agree, disagree, no stance                                                                                                                                             & \url{https://github.com/ucinlp/covid19-data}                                \\\\
SciFact~\cite{Wadden2020-kg}                                                 & COVID-19            & 2                & expert-written claims, scientific abstracts                                                                   & English                                              & text                              & SUPPORTS, REFUTES, NOINFO                                                                                                                          & \url{https://github.com/allenai/scifact}                                         \\\\
HealthVer~\cite{Sarrouti2021-cq}                                              & COVID-19            & 2                & claims returned by a search engine, scientific articles                                                       & English                                              & text                              & SUPPORTS, REFUTES, NEUTRAL                                                                                                                         & \url{https://github.com/sarrouti/healthver}                                      \\\\
Check-COVID~\cite{Wang2023-mu}                                            & COVID-19            & 2                & health news, scientific articles                                                                              & English                                              & text                              & SUPPORT, REFUTE, NEI                                                                                                                               & \url{https://github.com/posuer/Check-COVID}                                      \\\\
CoVERT~\cite{Mohr2022-bp}                                                  & COVID-19            & 2                & Twitter/X (tweets), scientific articles                                                                       & English                                              & text                              & SUPPORTS, REFUTES, NEI                                                                                                                             & \url{https://www.ims.uni-stuttgart.de/data/bioclaim}                             \\\\
ReCOVery~\cite{Zhou2020-mf}                                              & COVID-19            & 2                & health news, Twitter/X (tweets)                                                                               & English                                              & text, image                       & reliable, unreliable                                                                                                                               & \url{https://github.com/apurvamulay/ReCOVery}                                    \\\\
COVID-rumor~\cite{Cheng2021-wh}                                        & COVID-19            & 2                & health news, Twitter/X (tweets)                                                                               & English                                              & text                              & True, False, Unverified                                                                                                                            & \url{https://github.com/MickeysClubhouse/COVID-19-rumor-dataset}                 \\\\
COVID-19 Disinfo~\cite{Alam2021-cb}                                        & COVID-19            & 2                & Twitter/X (tweets)                                                                              & Arabic, English, Dutch, Bulgarian                                              & text                              & binary and multi-class                                                                                                                            & \url{https://github.com/firojalam/COVID-19-disinformation}                 \\\\
ArCOV19-Rumors~\cite{Haouari2021-lf}                                         & COVID-19            & 2                & Twitter/X (tweets)                                                                                            & Arabic                                               & text                              & True, False, Other                                                                                                                                 & \url{https://gitlab.com/bigirqu/ArCOV-19/-/tree/master/ArCOV19-Rumors}           \\\\
CHECKED~\cite{Yang2021-iq}                                                 & COVID-19            & 2                & Weibo                                                                                                         & Chinese                                              & text, image, video                & Real, Fake                                                                                                                                         & \url{https://github.com/cyang03/CHECKED}                                         \\\\
MM-COVID~\cite{Li2020-jn}                                              & COVID-19            & 2                & Twitter/X (tweets), fact-checking websites                                                                    & English, Spanish, Portuguese, Hindi, French, Italian & text, image                       & Real, Fake                                                                                                                                         & \url{https://github.com/bigheiniu/MM-COVID}                                      \\\\
MMCoVaR~\cite{Chen2021-vp}                                                 & COVID-19            & 2                & health news, Twitter/X (tweets)                                                                               & English                                              & text, image, temporal information & Support, Refute, Not Enough Information                                                                                                            & \url{https://github.com/InfintyLab/MMCoVaR}                                      \\\\
CoAID~\cite{Cui2020-uv}                                                 & COVID-19            & 2                & social media, fact-checking websites                                                                          & English                                              & text                              & True, Fake                                                                                                                                         & \url{https://github.com/cuilimeng/CoAID}                                         \\\\
ANTi-Vax~\cite{Hayawi2022-wz}                                              & COVID-19            & 2                & Twitter/X (tweets)                                                                                            & English                                              & text                              & Misinformation, General Vaccine-Related Tweets                                                                                                     & \url{https://github.com/sakibsh/ANTiVax}                                         \\\\
Infodemic2019~\cite{Luo2021-lq}                                         & COVID-19            & 2                & Weibo, WeChat mini-app                                                                                        & Chinese                                              & text                              & Questionable, False, True                                                                                                                          & \url{https://www.dropbox.com/sh/praltzebemotd2r/AABmc1IxaKG\_uZnEUN5beJFwa?dl=0} \\\\
MisinfoCorrect~\cite{He2023-du} & COVID-19 & 2 & Twitter/X (tweets and responses) & English & text & Polite, Neutral, Rude & \url{https://github.com/claws-lab/MisinfoCorrect} \\\\
LimeSoda~\cite{Payoungkhamdee2021-wj}                                               & general medical     & 2                & official healthcare departments, health news, online articles, e-commerce platforms, web boards, social media & Thai                                                 & text                              & Fact, Fake, Undefined                                                                                                                              & \url{https://github.com/byinth/LimeSoda}                                         \\\\
MuMiN~\cite{Nielsen2022-fs}                                                  & general medical     & 2                & Twitter/X (tweets), online articles, fact-checking websites                                                   & 41 languages                                         & text, image                       & factual, misinformation                                                                                                                            & \url{https://mumin-dataset.github.io/}                                           \\\\
Med-Fact~\cite{Tan2023-oi}                                                & general medical     & 2                & multiple-choice question-answering datasets                                                                   & English                                              & text                              & SUPPORTED, REFUTED, NOT ENOUGH INFO                                                                                                                & \url{https://github.com/taneset/Multi2Claim}                                     \\\\
HealthFC~\cite{Vladika2024-wl}                                               & general medical     & 2                & online health inquiries, clinical trials and systematic reviews                                               & English, German                                      & text                              & Supported, Refuted, NEI                                                                                                                            & \url{https://github.com/jvladika/HealthFC}                                       \\\\
BEAR-FACT~\cite{Wuehrl2024-ro}                                              & general medical     & 2                & Twitter/X (tweets), scientific articles                                                                       & English                                              & text                              & SUPPORTED, PARTIALLY SUPPORTED, REFUTED, PARTIALLY REFUTED, UNVERIFIABLE                                                                           & \url{https://www.ims.uni-stuttgart.de/data/bioclaim}                             \\\\
PubHealth~\cite{Kotonya2020-zy}                                              & general medical     & 2                & claims and cited sources from fact-checking and news websites, explanations by journalists                    & English                                              & text                              & TRUE, FALSE, MIXTURE, UNPROVEN                                                                                                                     & \url{https://github.com/neemakot/Health-Fact-Checking}                           \\\\
PubHealthTab~\cite{Akhtar2022-zj}                                           & general medical     & 2                & claim-table pairs from online articles                                                                        & English                                              & text                              & SUPPORTS, REFUTES, NEI                                                                                                                             & \url{https://github.com/mubasharaak/PubHealthTab}                                \\\\
Monant~\cite{Srba2022-cj}                                               & general medical     & 2                & health news, fact-checking websites                                                                           & English                                              & text                              & Supporting, Contradicting, Neutral                                                                                                                 & \url{https://github.com/kinit-sk/medical-misinformation-dataset}                 \\\\
Med-MMHL~\cite{Sun2023-jn}                                              & general medical     & 2, 3 & health news, Twitter/X (tweets), LLM-generated text                                                           & English                                              & text, image                       & Real, Fake                                                                                                                                         & \url{https://github.com/styxsys0927/Med-MMHL}                                    \\\\
Med-HALT~\cite{Pal2023-nm}                                              & general medical     & 3                 & multiple-choice questions, PubMed abstracts, LLM-generated text                                               & English                                              & text                              & binary                                                                                                                                             & \url{https://medhalt.github.io/}                                                 \\\\
Med-HallMark~\cite{Chen2024-aa}                                           & general medical     & 3                 & LVLM-generated text                                                                                           & English                                              & text, image                       & Catastrophic Hallucination, Critical Hallucination, Attribute Hallucination, Prompt-induced Hallucination, Minor Hallucination, Correct Statements & \url{https://github.com/ydk122024/Med-HallMark}                                  \\\\
MedVH~\cite{Gu2024-vy}                                                  & general medical     & 3                 & LVLM-generated text                                                                                           & English                                              & text, image                       & Wrongful Image, None of the Above, Clinically Incorrect Questions, False Confidence Justification, General Report Generation                       & \url{https://github.com/dongzizhu/MedVH}                                                 \\

\end{longtable}
\normalsize

\section{Errors}
\label{sec:errors}

Thirteen error-related articles were included in this review. Table \ref{tab:error} provides an overview of these articles. Most of the articles address general medical errors or medication errors, while a few focus on specific issues such as sedation in endoscopy~\cite{Shen2021-jw} and radiation oncology~\cite{Ganguly2023-gu}, as detailed in the ``Topic" column in Table 2. 

The types of source documents utilized in error-related articles can be classified into two primary categories (see Table \ref{tab:error_s1} of the Supplementary Information). The first category consists of incident reports~\cite{Wong2020-fm, Harkanen2020-ag}, patient safety reports~\cite{Boxley2023-hh, Eskildsen2020-yj}, and error reports~\cite{Ganguly2023-gu}, which typically describe clinical error processes directly. Technically, these texts should not be classified as inaccurate, as they document medical errors during patient care without inherently causing misunderstanding or inaccuracies. We selected studies focusing on this type of text because they offer a comprehensive understanding of medical error categories and assist in identifying potential inaccuracies within unstructured clinical text. Such documents are commonly employed in rule-based text mining and error classification tasks.

The second category includes authoritative texts (e.g., Wikipedia articles~\cite{Bravo-Candel2021-ko}, PubMed abstracts~\cite{Lee2022-hk}) and clinical documents (e.g., clinical notes~\cite{Valiev2024-vb, Toma2024-io, Gundabathula2024-vl}, pathology reports~\cite{Lee2022-hk}, prescriber directions~\cite{Pais2024-di}) that have been manually modified to introduce errors. These errors are introduced either by altering spelling or by changing the meaning of sentences through the substitution of key terms related to diagnosis, management, and treatment. The former is typically used in spelling correction tasks~\cite{Bravo-Candel2021-ko, Lee2022-hk}, while the latter is mainly applied in error detection and correction within clinical text~\cite{Valiev2024-vb, Toma2024-io, Gundabathula2024-vl}. Due to the sensitive nature of patient information and privacy concerns, most of the clinical datasets are not publicly available. 

\begin{table}[t]
\tiny
\centering
\caption{Overview of NLP research about medical errors (Task: 1 = error detection, 2 = error correction, 3 = others)}
\label{tab:error}
\begin{tabularx}{\textwidth}{@{}lXp{5em}p{10em}Xp{10em}p{10em}p{5em}@{}}
\toprule
Ref. & Topic & Task & Document type & Dataset & Method & Metrics & Factuality evaluation \\
\midrule
Wong et al.~\cite{Wong2020-fm}        & general medical           & 1           & incident reports                                                        & internal dataset                & DNN, logistic regression, support vector machines, decision trees & sensitivity, specificity, F1, accuracy, AUC                            & automatic                       \\
Boxley et al.~\cite{Boxley2023-hh}     & general medical           & 1           & patient safety event reports                                            & internal dataset                & logistic regression, elastic net, XGBoost                         & accuracy, precision, recall, specificity, F1, AUC-ROC, PR-ROC          & automatic                       \\
Eskildsen et al.~\cite{Eskildsen2020-yj}    & medication administration    & 1   & individual case safety reports                                          & internal dataset                & I2E text mining, CPR text mining                                  & precision, recall                                                      & automatic                       \\
Ganguly et al.~\cite{Ganguly2023-gu}     & radiation oncology        & 1           & error reports of radiation oncology                                     & internal dataset                & TF-IDF, LSA, SVM, MLP, CNN                                        & accuracy, precision, recall, F1                                        & automatic                       \\
Valiev et al.~\cite{Valiev2024-vb}       & general medical           & 1, 2                      & medical question-answering text and clinical notes with injected errors & MEDEC                & GPT-3.5, GPT-4                                                    & accuracy, ROUGE, BERTScore, BLEURT, AggregateComposite, AggregateScore & automatic                       \\
Toma et al.~\cite{Toma2024-io}       & general medical           & 1, 2                      & medical question-answering text and clinical notes with injected errors & MEDEC                & GPT-3.5, GPT-4, DSPy framework                                    & accuracy, ROUGE, BERTScore, BLEURT, AggregateComposite, AggregateScore & automatic                       \\
Gundabathula et al.~\cite{Gundabathula2024-vl} & general medical           & 1, 2 & medical question-answering text and clinical notes with injected errors & MEDEC                & GPT-3.5, GPT-4, Claude-3 Opus                                     & accuracy, ROUGE, BERTScore, BLEURT, AggregateComposite, AggregateScore & automatic                       \\
Pais et al.~\cite{Pais2024-di} & general medical & 1, 2                                     & prescriber directions from electronic prescriptions      & internal dataset & MEDIC, T5-FineTuned, Claude                       & BLEU, METEOR, near-miss ratios                                                & automatic, expert                       \\
Bravo-Candel et al.~\cite{Bravo-Candel2021-ko} & general medical & 2                                     & Wikipedia articles and clinical case reports with synthetic errors      & Spanish real word error dataset & seq2seq (RNN, transformer), GloVe, Word2Vec                       & precision, recall, F0.5                                                & automatic                       \\
Lee et al.~\cite{Lee2022-hk}          & general medical           & 2                                     & PubMed abstracts and surgical pathologic records with synthetic errors  & internal dataset                & MLM                                                               & precision, recall, F1                                                  & automatic                       \\
Härkänen et al.~\cite{Harkanen2020-ag}     & medication administration & 3: rule-based text mining                                         & incident reports                                                        & internal dataset                & SAS Text Miner                                                    & Fisher’s exact test                                                    & automatic, expert               \\
Shen et al.~\cite{Shen2021-jw}         & sedation in endoscopy     & 3: rule-based text mining          & historical endoscopy records                                            & internal dataset                & heuristic checks, keyword recognition                             & precision, recall, specificity, negative predictive value              & automatic, expert            \\   
Tavabi et al.~\cite{Tavabi2024-vz}      & procedural terminology    & 3: text classification           & operative notes                                                         & internal dataset                & TF-IDF, Doc2Vec, BERT                                             & accuracy, sensitivity, specificity, AUROC                              & automatic                       \\
\bottomrule
\end{tabularx}
\end{table}

\subsection{Error detection}

Error detection refers to identifying inaccuracies in data, often through classification tasks. This includes binary classification to detect the presence of errors, and multi-class or multi-label classification to categorize different types of errors. LLMs have also been utilized to integrate error classification into their reasoning processes to improve performance in subsequent correction tasks~\cite{Gundabathula2024-vl}. Common metrics used for error detection tasks include accuracy, precision, recall, F1-score (F1), and area under the curve (AUC). Some papers prefer to use sensitivity (recall) and specificity (negative predictive value) to describe the ability to detect true positives and true negatives, respectively~\cite{Wong2020-fm, Tavabi2024-vz}. None of our reviewed studies included expert evaluations for error detection tasks.

Binary classification tasks aim to identify the presence of a specific type of error. Eskildsen et al.~\cite{Eskildsen2020-yj} explored two text mining methods alongside the traditional Safety Surveillance Advisor (SSA) method to detect medication errors in individual case safety reports (ICSRs). The study focused on patients extracting insulin from prefilled pens or cartridges using a syringe, an action identified by the European Medicines Agency in 2017 as a medication error~\cite{European-Medicines-Agency2017-iq}. The dataset consisted of 154,209 ICSRs from Novo Nordisk’s safety database (1987–2018), with 2,533 cases manually annotated for testing. These reports included narratives coded with MedDRA~\cite{Brown1999-dz} terms related to device or medication use errors. While the three methods demonstrated relatively high recall, ranging from 0.785 to 0.904, precision was notably low, ranging from 0.016 to 0.034, due to high false positive rates. 

Multi-class classification tasks categorize errors into one of several predefined error groups. Ganguly et al.~\cite{Ganguly2023-gu} developed automated error-labeling models to classify errors in radiation oncology. Their study utilized 1,121 error reports from a radiation oncology center's Medical Error Reduction Program (MERP) database\footnote{\url{https://radphysics.com/}}. The dataset included free-text descriptions and event category labels, grouped into four broad categories: Administrative, Standards, Treatment, and Treatment Preparation. Key methods in this study included Linear Support Vector Machine (SVM), Multilayer Perceptron (MLP), and Convolutional Neural Networks (CNN), with features derived from Term Frequency-Inverse Document Frequency (TF-IDF) and reduced through Latent Semantic Analysis (LSA). The performance of the SVM and MLP models was robust, with weighted F1-scores ranging from 0.874 to 0.998. However, the CNN model underperformed, likely due to the limited size of the dataset. This study highlighted the effectiveness of models in detecting human labeling errors and reducing heuristic bias - the tendency of human reporters to rely on subjective judgment or limited perspectives when categorizing errors, which can lead to inconsistent labeling. 

Multi-label classification tasks assign multiple error categories to a single instance. Wong et al.~\cite{Wong2020-fm} applied deep neural network (DNN) models (feedforward artificial neural networks with varying architectures consisting of 1-5 hidden layers) to classify medication administration errors, focusing on wrong patient, wrong drug, wrong time, wrong dose, and wrong route. The dataset consisted of 574 incident reports collected from the Hong Kong Hospital Authority’s Advanced Incident Reporting System (AIRS) over 2011–2014, with structured and unstructured free-text fields describing medication errors and near-misses. DNN models achieved high performance, with an average accuracy of 0.94 and an AUC of 0.911 across all five categories. Boxley et al.~\cite{Boxley2023-hh} employed logistic regression, elastic net, and XGBoost models to classify medication errors in patient safety event (PSE) reports using eight categories adapted from the NCC MERP taxonomy~\cite{National-Coordinating-Council-for-Medication-Error-Reporting-and-Prevention-NCC-MERP-2001-zs}, such as wrong drug, wrong time, wrong dose, and monitoring errors. The dataset included 3,861 annotated PSE reports from a ten-hospital healthcare system, with structured fields and free-text narratives describing medication safety events. Among the models tested, XGBoost demonstrated the highest performance, with an average F1-score of 0.72 across categories. These error types and categories play a crucial role in addressing inaccurate information by identifying underlying patterns of medication errors, ultimately enhancing safety processes and reducing the likelihood of similar mistakes in the future.

LLM-based classification methods integrate error detection into the inference process to enhance downstream tasks. Gundabathula et al.~\cite{Gundabathula2024-vl} categorized errors into domains (e.g., medications, medical conditions, clinical procedures) and included this structure in model prompts. This guided the model through a CoT reasoning process, which improved the model’s accuracy and ensured more precise and explainable results. In GPT-3.5, classification accuracy increased from 48.75$\%$ to 58.44$\%$, and span identification accuracy from 22.5$\%$ to 38.55$\%$. Using GPT-4 further improved performance, reaching 63.07$\%$ and 58.17$\%$ for those tasks, respectively. It also helped reduce hallucination and enhance consistency.

\subsection{Error correction}

In the reviewed articles, error correction primarily includes spelling correction and contextual error correction. Spelling correction focuses on typographical and lexical errors, while contextual error correction addresses issues such as incorrect diagnoses, treatments, or medication instructions derived from broader contextual information. In spelling correction tasks, the most common evaluation metrics are precision, recall, and F1. Bravo-Candel et al.~\cite{Bravo-Candel2021-ko} used F0.5 instead of F1, giving more weight to precision. This choice was made because in spelling correction tasks, false positives are often undesirable, making models with higher precision (fewer false positives) preferable. In contextual error detection tasks, accuracy was used to evaluate the correct identification of errors (error flagging) and the correct detection of sentences containing errors~\cite{Ben-Abacha2024-yk}. In contextual error correction tasks, ROUGE, BERTscore, and BLEURT were commonly used. ROUGE measured unigram overlap between generated and reference text, commonly used for sentence correction~\cite{Lin2004-vu}. BERTScore used contextual embeddings to assess semantic similarity between generated and reference sentences~\cite{Zhang2019-me}, while BLEURT employed machine-learned metrics trained on human ratings for deeper quality evaluation~\cite{Sellam2020-sc}. AggregateScore combined multiple metrics to provide a balanced measure of performance across different dimensions, and was used for ranking in MEDIQA-CORR 2024~\cite{Ben-Abacha2024-yk}. Only one paper involved clinical expert evaluating the system's outputs to identify potential near-miss events and validate the safety and accuracy of medication directions~\cite{Pais2024-di}.

Two papers specifically focused on spelling correction. Bravo-Candel et al.~\cite{Bravo-Candel2021-ko} used Seq2Seq neural machine translation models to correct real-word errors  in Spanish clinical texts. The study used two datasets: the Wikicorpus, comprising over 611 million words from Spanish Wikipedia articles, and the medicine corpus, a smaller dataset of approximately 5,750 clinical cases with around 2 million words from three Spanish clinical corpora (CodiEsp~\cite{Miranda-Escalada2020-kz}, MEDDOCAN~\cite{Marimon2019-jm}, SPACCC~\cite{Intxaurrondo2018-cb}). Errors were synthetically introduced in the sentences using predefined rules across six categories, such as grammatical gender and subject-verb concordance. The best performance was achieved with models trained on the medicine corpus, yielding an F0.5 score of 0.6498 with no pre-trained embeddings. The Spanish-language dataset used in this study is one of the few publicly available clinical datasets for error detection and correction. Lee et al.~\cite{Lee2022-hk} employed a Masked Language Model (MLM)-based approach to correct spelling errors in unstructured medical texts and enhance NER accuracy. The study utilized two datasets: the NCBI-disease corpus~\cite{Dogan2014-od} (793 PubMed abstracts annotated with disease mentions) and surgical pathology records (40,443 annotated lung cancer diagnostic records from the Asan Medical Center). Synthetic errors were introduced to mimic real-world typographical patterns. The MLM-based spelling correction achieved F1-scores of 0.72 (NCBI-disease) and 0.73 (surgical pathology records). For NER tasks, spelling correction significantly boosted F1-scores in surgical pathology records from 0.60 to 0.85. This study highlighted the potential of spelling correction models to mitigate data quality issues and improve the accuracy of downstream NLP tasks in clinical settings.

Contextual error correction has seen advancements through the application of LLMs for addressing complex, context-dependent inaccuracies. A highlight in this domain is the MEDIC system introduced by Pais et al.~\cite{Pais2024-di}, which focuses on preventing medication direction errors in online pharmacies by improving accuracy and standardization during the data entry phase. The study utilized 1.6 million single-line medication directions from Amazon Pharmacy, including raw prescriber directions and pharmacist-verified equivalents. MEDIC employs a three-stage process: pharmalexical normalization, which standardizes and corrects variations in medication terminology and formatting; AI-powered extraction using a fine-tuned DistilBERT model; and semantic assembly with safety guardrails informed by a medication catalog derived from RxNorm, OpenFDA, and Amazon Pharmacy data. Compared to T5-FineTuned and Claude, MEDIC reduced near-miss events by 33$\%$ during deployment, with a notable improvement in suggestion adoption rates and a reduction in post-adoption edits. 

The MEDIQA-CORR 2024 shared task focuses on error detection and contextual error correction in clinical text, encompassing three subtasks: binary classification of texts with errors, identification of erroneous sentences, and generation of corrected text~\cite{Ben-Abacha2024-yk}. The dataset includes 3,848 clinical texts derived from two sources: the Microsoft (MS) collection, transformed from the MedQA~\cite{Jin2019-jf} dataset with manual error injections, and the University of Washington (UW) collection, containing de-identified clinical notes from UW Medical Center~\cite{Abacha2024-se}. The MS training set contains 2,189 texts, with validation sets for both MS (574 texts) and UW (160 texts), and the test set combines texts from both sources, with 597 texts from MS and 328 texts from UW. Errors were annotated to simulate real-world scenarios, covering categories like diagnoses, treatments, and pharmacotherapy. Evaluation metrics included ROUGE-1, BERTScore, and BLEURT, with the aggregate score serving as the main ranking criterion. Top 3 ranked papers out of the 17 teams that participated in MEDIQA-CORR 2024 are included in this review. Valiev et al.~\cite{Valiev2024-vb} employed a multi-component approach combining named entity recognition (NER), knowledge graph integration using MeSH, and an ensemble of outputs from multiple LLMs (GPT-3.5, GPT-4, and Claude). Their system achieved a BERTScore of 0.806 and aggregate score of 0.781, ranking third overall. Gundabathula et al.~\cite{Gundabathula2024-vl} adopted a self-consistency strategy and ensemble approaches to enhance robustness in prompt-based in-context learning. By combining GPT-4 and Claude-3 Opus outputs, their system achieved an aggregate score of 0.787, ranking second out of 17 teams. Toma et al.~\cite{Toma2024-io} used a retrieval-based approach leveraging MedQA datasets, optimized prompts, and the DSPy~\cite{Khattab2023-vl} framework for few-shot learning to handle subtle errors in the MS dataset and more explicit errors in the UW dataset. Their system achieved an aggregate correction score of 0.789, and ranked first in the MEDIQA-CORR task. However, they queried the MedQA dataset, potentially leading to test data leakage since MedQA was used to construct the MS dataset. This overlap raises concerns that using external datasets for retrieval-based methods may lead to data leakage, potentially inflating performance metrics and affecting the generalizability of the approaches~\cite{Fu2024-uu}.

\subsection{Others}

Apart from error detection and correction, earlier error-related work employed rule-based text mining methods to address inaccurate information in clinical contexts. For instance, Härkänen et al.~\cite{Harkanen2020-ag} employed SAS Text Miner (a text analysis tool using the bag-of-words method on the SAS Enterprise Miner platform) to analyze incident reports and investigate the relationship between staffing-related word triggers (e.g., ``short staffing" and ``workload") and error types, with a particular focus on medication administration errors. This study included manual labeling of medical harm, illustrating the degree of harm associated with different triggers. Shen et al.~\cite{Shen2021-jw} used heuristic checks and keyword recognition on historical endoscopy records to predict appropriate sedation strategies. This approach involved checking case-insensitive matches and correcting term discrepancies across records, which helped correct inaccurate information in records and erroneous sedation orders. The inclusion of an endoscopy triage nurse for manual review further enhanced the reliability of the automated system. 

Tavabi et al.~\cite{Tavabi2024-vz} highlighted a critical issue in codification errors related to gold standard labels while evaluating TF-IDF, Doc2Vec, and BERT models for assigning current procedural terminology (CPT) codes from operative notes. The study, focused on musculoskeletal procedures, used a dataset of 44,002 operative notes annotated with the 100 most common CPT codes. TF-IDF demonstrated superior performance with an AUROC of 0.96 and accuracy of 0.97, outperforming both Doc2Vec and BERT, likely due to its robustness to data sparsity and noise in clinical notes. Importantly, the study revealed discrepancies in gold standard CPT assignments during experiments, with NLP models flagging potentially mislabeled instances and correcting errors in the provided ground truth in some cases. This underscores the need for refining data labeling processes to enhance the reliability of automated codification systems.

Several other studies focused on reducing cognitive overload among healthcare providers, which indirectly helps in reducing medically inaccurate information~\cite{Gao2022-va,Gao2023-ue,Dymek2021-ns,Murray2021-nx,Su2020-jq}. These papers, although not included in this review, highlight the growing role of NLP in supporting healthcare providers by improving diagnostic accuracy and reducing cognitive biases. 

\section{Misinformation}
\label{sec:misinformation}

A total of 27 articles related to misinformation were included in this study. We provided an overview of these articles in Table \ref{tab:misinformation}. 14 out of 27 articles are related to COVID-19. The other articles are on topics such as the HPV vaccine~\cite{Du2021-xy,Chin2020-pr}, dermatology~\cite{Sager2021-xh}, cardiology, gynecology, psychiatry, and pediatrics~\cite{Nabozny2023-bg}. Two main tasks in addressing misinformation are (1) misinformation detection and (2) misinformation correction.

Misinformation detection encompasses two types of tasks: direct misinformation identification and fact-checking and claim verification. Direct misinformation identification focuses on directly classifying content as real or fake without requiring external evidence for evaluation. The input for this task often consists of social media posts (e.g., Twitter/X, Reddit)~\cite{Alam2021-cb,Haupt2024-ir,Sager2021-xh,Du2021-xy} or health news articles~\cite{Zuo2021-tu}. However, due to privacy concerns associated with social media platforms and health news websites, most datasets for direct misinformation identification are not publicly available. Fact-checking and claim verification involves evaluating specific claims to determine their veracity by comparing them against external evidence. Claims are often sourced from fact-checking websites (e.g., Science Feedback, FactCheck.org, Snopes, PolitiFact)~\cite{Kotonya2020-zy,Akhtar2022-zj}, health news articles~\cite{Wang2023-mu,Akhtar2022-zj,Deka2022-vj}, social media platforms~\cite{Mohr2022-bp,Martin2022-js,Wuhrl2022-em,Wuehrl2024-ro}, or medical QA datasets~\cite{Tan2023-oi}. Unlike direct misinformation identification, fact-checking and claim verification requires additional input in the form of evidence, which is often derived from trusted sources such as PubMed abstracts~\cite{Wadden2020-kg,Liu2020-bc}, research papers~\cite{Sarrouti2021-cq,Mohr2022-bp,Wuhrl2022-em,Wang2023-mu,Deka2022-vj,Wuehrl2024-ro}, clinical trials and systematic reviews~\cite{Vladika2024-wl}. Fact-checking and claim verification benefits from the availability of several publicly accessible datasets.

Misinformation correction is less explored compared to detection, with only two studies in this review addressing it~\cite{He2023-du,Yue2024-at}. Both studies focused on generating polite, evidence-backed responses using social media posts and academic articles as sources of misinformation and supporting evidence. The two papers either released datasets or built on publicly available datasets, but further datasets are still needed to support advancements in this area.

\begin{table}[H]
\tiny
\centering
\caption{Overview of NLP research about medical misinformation (Task: 1 = misinformation detection, 2 = misinformation correction, 3 = others, 1: (a) = direct misinformation identification, 1: (b) = fact-checking and claim verification )}
\label{tab:misinformation}
\begin{tabularx}{\textwidth}{@{}XXp{8em}p{10em}Xp{10em}p{10em}p{5em}@{}}
\toprule
Ref. & Topic & Task & Document type & Dataset & Method & Metrics & Factuality evaluation \\
\midrule
Alam et al.~\cite{Alam2021-cb}   & COVID-19                                           & 1: (a)                                                                  & Twitter/X (tweets)                                                & COVID-19 Disinfo             & BERT, RoBERTa, XLM-R, AraBERT, BERTje                                  & accuracy, precision, recall, F1                                   & automatic                                              \\
Haupt et al.~\cite{Haupt2024-ir}   & COVID-19                                           & 1: (a)                                                                  & Twitter/X (tweets)                                                & internal dataset            & GPT-3.5-turbo                                  & accuracy                                  & automatic                                              \\
Sager et al.~\cite{Sager2021-xh}   & dermatology (tanning and essential oil)                                           & 1: (a)                                                          & Reddit posts                                               & internal dataset              & LR, BERT, XLNet                                  & accuracy                           & automatic                                              \\
Zuo et al.~\cite{Zuo2021-tu}      & general medical                                    & 1: (a)                                                              & health news                                                                                & internal dataset    & SVM, GB, BERT, XLNet, RoBERTa, ALBERT, DistilBERT, Longformer             & precision, recall, F1                                             & automatic                                              \\
Du et al.~\cite{Du2021-xy}      & HPV vaccine                                        & 1: (a)                                                              & Reddit posts                                                                               & internal dataset    & SVM, LR, extremely randomized trees, CNN, RNN, BTM                        & precision, recall, F1, AUC                                        & automatic                                              \\
Garbarino et al.~\cite{Garbarino2024-ao}	& sleep health	& 1: (a)	& sleep-related myths compiled from literature	& internal dataset	& ChatGPT-4, Google Bard	& ICC coefficient	& automatic  \\
Wang et al.~\cite{Wang2021-mc} & vaccination	& 1: (a)	& Instagram posts	& internal dataset & 	RNN, VGG19	& accuracy, precision, recall, F1	& automatic \\
Wadden et al.~\cite{Wadden2020-kg}	& COVID-19 	& 	1: (b)		& expert-written claims, scientific abstracts		& SCIFACT	& 	VeriSci, SciBERT, BioMedRoBERTa, RoBERTa	& 	accuracy, precision, recall, F1		& automatic\\
Liu et al.~\cite{Liu2020-bc}      & COVID-19                                           & 1: (b)                                                                          & claims, scientific abstracts, Wikipedia documents                                          & SCIFACT, FEVER      & SciKGAT, SciBERT, RoBERTa                                                 & precision, recall, F1                                             & automatic                                              \\
Sarrouti et al.~\cite{Sarrouti2021-cq} & COVID-19                                           & 1: (b)                                      & claims returned by a search engine, scientific articles                                    & HEALTHVER           & BM25, T5, BERT, SciBERT, BioBERT                                          & P@10, R@10, NDCG@10, accuracy, precision, recall, F1              & automatic                                              \\
Mohr et al.~\cite{Mohr2022-bp}     & COVID-19                                           & 1: (b)                & Twitter/X (tweets), scientific articles                                                      & CoVERT              & BERT, BioBERT, scispaCy, MLP                                              & Acc@1, Acc@5, precision, recall, F1                               & automatic                                              \\
Martín et al.~\cite{Martin2022-js}  & COVID-19                                           & 1: (b)                                                                       & Twitter/X (tweets), fact-checked information                                                 & NLI19-SP            & XLM-RoBERTa, BERT                                                         & precision, recall, F1                                             & automatic                                              \\
Wührl et al.~\cite{Wuhrl2022-em}    & COVID-19                                           & 1: (b)        & Twitter/X (tweets), scientific articles                                                      & CoVERT              & MultiVerS                                                                 & precision, recall, F1                                             & automatic                                              \\
Wang et al.~\cite{Wang2023-mu}    & COVID-19                                           & 1: (b)                                                                & health news, scientific articles                                                           & Check-COVID Dataset & RoBERTa, BM25, GPT-3.5                                                    & accuracy, precision, recall, F1                                   & automatic                                              \\
Kotonya et al.~\cite{Kotonya2020-zy}  & general medical                                    & 1: (b)                                   & claims and cited sources from fact-checking and news websites, explanations by journalists & PubHealth           & BERT, SciBERT, BioBERT, S-BERT                                            & accuracy, precision, recall, F1, ROUGE                            & automatic, expert                                      \\
Akhtar et al.~\cite{Akhtar2022-zj}  & general medical                                    & 1: (b)                                                            & claim-table pairs from online articles                                                     & PubHealthTab        & RoBERTa, BERT, BioBERT, ClinicalBERT, BlueBERT, ALBERT, TAPAS, T5         & F1                                                                & automatic                                              \\
Deka et al.~\cite{Deka2022-vj}    & general medical                                    & 1: (b)                                 & online articles, scientific articles                                                       & internal dataset    & TextRank, S-BERT, scispaCy, SapBERT, K-Means Clustering, BioBERT, RoBERTa & precision, recall, F1                                             & automatic                                              \\
Tan et al.~\cite{Tan2023-oi}     & general medical                                    & 1: (b)                                           & multiple-choice question-answering datasets                                                & Med-Fact            & BART, BERT, DeBERTa, SciBERT, Longformer, BioBERT                         & weighted-F1, fluency, contextually, faithfulness, challenge level & automatic, expert                                      \\
Wührl et al.~\cite{Wuehrl2024-ro}   & general medical                                    & 1: (b) & Twitter/X (tweets), scientific articles                                                      & BEAR-FACT Corpus    & RoBERTa                                                                   & precision, recall, F1                                             & automatic
\\
Vladika et al.~\cite{Vladika2024-wl}  & general medical                                    & 1: (b)                                                                & online health inquiries, clinical trials and systematic reviews                            & HealthFC            & BERT, BioBERT, DeBERTa                                                    & precision, recall, F1                                             & automatic                                              \\
He et al.~\cite{He2023-du}     & COVID-19                                   & 2                                                                & Twitter/X (tweets and responses)                                                             & MisinfoCorrect    & BERT, RoBERTa, FC-GEN, DialoGPT, Seq2Seq, BART, Partner, GPT-2            & politeness, refutation, evidence support, fluency, relevance      & automatic                                              \\
Yue et al.~\cite{Yue2024-at} & COVID-19 & 2 & claims, scientific articles & Check-COVID, CORD, LitCovid & RARG & NDCG, recall, refutation, factuality, politeness, claim relevance, evidence relevance & automatic \\
Nabożny et al.~\cite{Nabozny2023-bg}  & cardiology, gynecology, psychiatry, and pediatrics & 3: credibility classification                                                                                  & online articles                                                                            & internal dataset    & LR, MLP, GB, BioBERT, LIME                                                & precision, recall, F1                                             & automatic, expert                                      \\
Zhou et al.~\cite{Zhou2020-mf} & 	COVID-19 & 	3: credibility classification & 	health news, Twitter/X (tweets) & 	ReCOVery & 	LIWC, RST, Text-CNN, SAFE	 & precision, recall, F1	 & automatic \\
Cheng et al.~\cite{Cheng2021-vm}   & COVID-19                                           & 3: misinformation network evolution analysis, misinformation central nodes prediction                          & Twitter/X (tweets)                                                                           & USC Melady Lab      & DNN, BERT embeddings                                                      & accuracy, AUROC, degree, closeness, betweenness                   & automatic                                              \\
Chin et al.~\cite{Chin2020-pr}     & HPV vaccine                                        & 3: psycholinguistics analysis, sentiment analysis, semantic representations                                    & online articles, news, and blogs                                                           & internal dataset    & Coh-Metrix, NLTK, Word2Vec, FastText, LSI                                 & narrativity, familiarity, semantic distance                       & automatic                                              \\

Hossain et al.~\cite{Hossain2020-bc} & 	COVID-19 & 	3: stance detection & 	Twitter/X (tweets) & 	COVID-Lies & 	BM25, BERTScore, SBERT & 	Hits@k, MRR, precision, recall, F1 & 	automatic \\

\bottomrule
\end{tabularx}
\end{table}

\subsection{Misinformation detection}

Although both error detection and misinformation detection involve identifying inaccuracies, their scope and methodologies differ in several ways. Error detection typically focuses on data from clinical or organizational systems (e.g., medical reports or patient records) where inaccuracies often arise from human or system mistakes, and it relies on fixed error categories (like ``wrong dose'' in medication records). In contrast, misinformation detection is frequently applied to open-domain content (e.g., social media posts, news articles), where the falsehood may be context-dependent or intentionally deceptive. Error detection usually benefits from standardized taxonomies and well-defined benchmarks, while misinformation detection may require more extensive factual validation and claim verification. Additionally, the level of domain expertise needed can vary: error detection may draw on established clinical guidelines or known protocols, whereas misinformation detection may depend on extensive domain knowledge (e.g., distinguishing partially correct statements from fully incorrect ones). Despite these differences, both tasks frequently utilize classification models and share core evaluation metrics such as accuracy, precision, recall, F1, and sometimes expert validation. Ultimately, error detection aims to correct mistakes in data and clinical workflows, whereas misinformation detection focuses on stemming the spread of deceptive or misleading content. Both approaches increasingly leverage LLMs to improve classification accuracy and interpretability.

\subsubsection{Direct misinformation identification}

The COVID-19 outbreak in 2020 sparked a surge in research on direct misinformation identification, resulting in numerous articles published between 2020 and 2022~\cite{Kotseva2023-fx}. Most of these studies used classification-based methods, in which machine learning or deep learning models were trained on internal datasets to classify content. The typical input for these models includes textual data from social media posts or health news articles, while the output is usually a binary or multi-class label indicating the veracity of the content~\cite{Zuo2021-tu,Du2021-xy,Sager2021-xh,Alam2021-cb}. In addition to text-based approaches, some studies adopted multimodal techniques by incorporating image or video features as input alongside textual data to improve detection accuracy~\cite{Wang2021-mc}. However, one major challenge in direct misinformation identification is developing a universal model that works across all topics. Simple classification methods often struggle with the context-specific nature of health misinformation, where truthfulness depends on detailed medical knowledge~\cite{Schlicht2023-ke}. Moreover, binary classification models that label information as real or fake can fail to capture the nuances of misleading but partially correct statements~\cite{Abdali2025-ek}. More recently, the advent of LLMs has introduced new possibilities for direct misinformation identification, including generative tasks where the models generate labels or explanations for the veracity of the input content~\cite{Haupt2024-ir,Garbarino2024-ao}. The key metrics used to evaluate models in direct misinformation identification are accuracy, precision, recall, and F1. One LLM-based study used intra-class correlation coefficient to evaluate the alignment between the outputs of LLMs and expert opinions~\cite{Garbarino2024-ao}. None of the reviewed studies included expert evaluations for direct misinformation identification tasks.

Text-based classification methods rely primarily on textual input and use traditional machine learning or deep learning models for binary or multi-class classification. Many studies classify text based on specific criteria or focus on identifying misinformation within particular topics. Zuo et al.~\cite{Zuo2021-tu} evaluated whether medical news articles meet a set of criteria to ensure the accuracy of health news. These ten criteria, developed collaboratively by healthcare journalists and medical professionals, included discussing costs, benefits, and harms of medical interventions, assessing evidence quality, and avoiding sensational language or disease-mongering. The study utilized a dataset of 1,119 medical news articles collected from Health News Review (website unavailable since 2022), comprising 740 news stories and 379 public relations releases. Each article was annotated according to these ten criteria. For experimentation, they focused on six key criteria that did not require highly specialized medical knowledge: costs of the intervention, quantification of benefits, quantification of harms, evidence quality, comparison with alternatives, and treatment availability. The study compared feature-based models (SVM and Gradient Boosting) with transformer-based models (BERT, RoBERTa~\cite{Liu2019-yy}, and Longformer~\cite{Beltagy2020-zq}). Gradient Boosting achieved the best F1-scores, exceeding 0.6 for four of the six selected criteria, while transformer models struggled due to data sparsity and article length exceeding token limits. Du et al.~\cite{Du2021-xy} identified misinformation in Reddit posts related to the HPV vaccine. They compiled 28,121 posts and manually labeled a subset of 2,200 posts to create a gold standard for evaluating models. Each post was annotated with one of binary labels: misinformation or nonmisinformation. The CNN model performed best, with an AUC of 0.7943 and an F1 score of 0.4925. Sager et al.~\cite{Sager2021-xh} focused on detecting misinformation in Reddit posts about tanning and essential oils in dermatology forums. Using Google BigQuery, they collected Reddit posts from January 2018 to August 2019 and filtered them based on keywords. The dataset included 2,608 posts, with 1,971 training instances and 221 test instances for essential oils, and 586 training instances and 66 test instances for tanning. Two medical students manually annotated the posts as either containing misinformation or not. The fine-tuned BERT model performed best, achieving 100$\%$ accuracy in detecting tanning-related misinformation and 99.56$\%$ accuracy for essential oils. Alam et al.~\cite{Alam2021-cb} addressed COVID-19 misinformation on Twitter/X by developing a large-scale, multilingual dataset of 16,000 manually annotated tweets in four languages (English, Arabic, Bulgarian, and Dutch). The annotations included dimensions such as factuality, harmfulness, verification need, and public interest. Results showed that RoBERTa achieved the best performance for English tweets (e.g., F1 of 0.964 for public interest binary classification, F1 of 0.856 for binary harmfulness classification). Similarly, XLM-R~\cite{Conneau2019-mm} performed best for other languages, achieving F1 scores ranging from 0.840 to 0.960. Additionally, this study demonstrated the benefits of multilingual and multitask learning. By combining data from four languages (English, Arabic, Bulgarian, and Dutch) and leveraging interrelated tasks, such as factuality assessment, check-worthiness, and societal harm detection, the authors improved classification performance for individual tasks. For instance, multitask learning significantly enhanced the performance on tasks like determining the need for fact-checking and assessing harm by using auxiliary tasks (e.g., factuality and public interest). This approach also addressed resource limitations in low-resource languages, as multilingual models like XLM-R outperformed monolingual ones in several cases, demonstrating the potential of cross-language knowledge transfer.

Multimodal methods leverage multiple types of input, such as text, images, and videos, to better identify and analyze misinformation in complex social media environments. Wang et al.~\cite{Wang2021-mc} developed a multimodal deep learning network to detect antivaccine messages on Instagram. The dataset, collected between 2016 and 2019, consists of over 30,000 Instagram posts evenly split between antivaccine and non-antivaccine content, annotated through a majority voting process among three trained annotators. The proposed model integrates features from images, text captions, hashtags, and Optical Character Recognition (OCR) results, utilizing a three-branch architecture with attention mechanisms to fuse multimodal information. An ensemble method further combines single-modal and multimodal outputs for improved accuracy. The model achieved an F1-score of 0.973, with text-based features contributing more significantly than visual data.

LLM-based approaches for direct misinformation identification leverage advanced contextual understanding, generative reasoning, and zero-shot learning capabilities of models. Haupt et al.~\cite{Haupt2024-ir} explored the impact of role-playing prompts on ChatGPT's accuracy in identifying COVID-19 misinformation, using a dataset of 36 tweets categorized into misinformation, unaligned sentiment, aligned sentiment, corrections, and neutral reporting. Each tweet was tested with 48 identity combinations (spanning political beliefs, education, locality, religiosity, and personality) and run 30 times to account for ChatGPT’s variability, resulting in 51,840 total responses. When no identities were included in prompts, ChatGPT achieved an average accuracy of 0.681; however, accuracy decreased to 0.293 when all identities were incorporated and further decreased to 0.192 when only political identities were included. This study highlighted challenges such as bias and inconsistency in ChatGPT's reasoning and emphasized the importance of prompt engineering and human oversight. Garbarino et al.~\cite{Garbarino2024-ao} evaluated the ability of ChatGPT-4 and Google Bard to debunk 20 common sleep-related myths, which were gathered from public sources. Annotations were provided by sleep medicine experts, detailing the falsity and public health significance of each myth. Both models were evaluated for their accuracy and alignment with expert assessments. ChatGPT-4 achieved an accuracy of 0.850 and a perfect positive predictive value of 100$\%$, demonstrating strong alignment with expert opinions, as reflected in an intra-class correlation coefficient of 0.83. In comparison, Google Bard outperformed ChatGPT-4 with an accuracy of 0.950.  Experts were directly involved in evaluating the AI outputs, ensuring professional oversight.

\subsubsection{Fact-checking and claim verification}

Fact-checking and claim verification is the process of determining the truthfulness of a claim by evaluating the alignment between a claim and relevant evidence~\cite{Panchendrarajan2024-nq}. A claim is a statement or assertion from healthcare-related sources, while evidence refers to supporting information from reliable scientific databases or factual records. The input for this task is a claim-evidence pair, and the output is a veracity label, which reflects the truthfulness of a claim or the alignment between the claim and evidence. Table \ref{tab:misinfo_s3} of the Supplementary Information shows several examples of claims and evidence in existing fact-checking datasets. This process involves five subtasks: (1) claim identification and extraction, (2) evidence retrieval, (3) evidence matching, (4) veracity prediction, and (5) validation and interpretation. The first three subtasks focus on corpus creation, while the last two are centered on building claim verification systems.\\

\textbf{Corpus Creation for Claim Verification}

Claim identification and extraction is the initial task of creating a fact-checking corpus involving the sourcing and formulation of claims from a variety of healthcare information sources. The following outlines several strategies employed to generate claims from the collected articles.  Wadden et al.~\cite{Wadden2020-kg} identified claims from citation sentences within scientific articles, which were then reformulated into atomic scientific claims by expert annotators. This study constructed the SCIFACT dataset, consisting of 1,409 claims paired with 5,183 abstracts from well-regarded journals, annotated with labels (SUPPORTS, REFUTES, or NOINFO) and rationales. Expert annotators also introduced negations to existing claims to make them refutable. Kotonya et al.~\cite{Kotonya2020-zy} used claims originating from fact-checking websites (e.g., Snopes\footnote{\url{https://www.snopes.com/}}, Politifact\footnote{\url{https://www.politifact.com/}}, FactCheck.org\footnote{\url{https://www.factcheck.org/}}) and news websites (e.g., Reuters\footnote{\url{https://www.reuters.com/}}, Associated Press\footnote{\url{https://apnews.com/}}), focusing on health-related articles addressing biomedical, health policy, and public health issues. The study introduced the PUBHEALTH dataset comprising 11,832 claims annotated with four veracity labels (true, false, mixture, unproven) and supplemented by journalist-curated gold-standard explanations to support the veracity assessments. Claims were processed and filtered based on a lexicon of public health terms to ensure relevance. Sarrouti et al.~\cite{Sarrouti2021-cq} retrieved claims from snippets generated by the Bing search engine in response to questions about COVID-19, focusing on naturally occurring information from the web. The study introduced the HEALTHVER dataset, which includes 1,855 claims and 738 associated evidence statements, resulting in 14,330 evidence-claim pairs annotated with SUPPORTS, REFUTES, and NEUTRAL labels. Mohr et al.~\cite{Mohr2022-bp} gathered claims from COVID-19-related tweets by using medical terms from the MeSH database, focusing on biomedical information shared about COVID-19 from January 2020 to June 2021. They further refined the dataset by selecting tweets that included causal relationships. The resulting CoVERT dataset consists of 300 annotated tweets containing biomedical claims, along with named entities (e.g., medical conditions, treatments) and relations (e.g., ``cause of"). Deka et al.\cite{Deka2022-vj} extracted claims from online health-related articles using TextRank\cite{Mihalcea2004-qi}, a graph-based ranking algorithm, to score and rank sentences according to their significance within the document. The authors assembled a dataset of 88 health-related articles from naturalnews.com, focusing on topics such as cancer and COVID-19. Each sentence in the dataset was manually annotated for claim relevance, with claims being identified based on their alignment with the article's heading. The approach employed unsupervised methods for claim extraction, utilizing semantic similarity calculations derived from pre-trained S-BERT embeddings. Tan et al.~\cite{Tan2023-oi} developed a pipeline to generate scientific claims from multiple-choice questions in scientific QA datasets, transforming them into declarative claims using a sequence-to-sequence model (BARTQA2D). The study introduced two datasets: (1) Med-Fact, with 150,000 claims in the biomedical domain, and (2) Gsci-Fact, with 32,000 claims in general science. The datasets feature a balanced distribution of claims classified as SUPPORTED, REFUTED, or NOT ENOUGH INFO. 

Evidence retrieval is the second step in fact-checking, involving the collection of relevant documents to support or refute a claim. In published literature, primary evidence sources included scientific articles (e.g., full papers~\cite{Kotonya2020-zy,Sarrouti2021-cq,Wang2023-mu}, PubMed abstracts~\cite{Wadden2020-kg,Liu2020-bc}, systematic reviews, and clinical trials~\cite{Vladika2024-wl}). Additionally, Akhtar et al.~\cite{Akhtar2022-zj} utilized web tables from over 300 websites linked to Wikipedia articles as evidence, offering structured data for claim verification. Once the evidence sources are gathered, it is essential to identify the most relevant documents. Techniques such as TF-IDF similarity were commonly employed for initial document retrieval~\cite{Wadden2020-kg,Liu2020-bc}, while more refined ranking methods included models like BM25 and Text-to-Text Transfer Transformer (T5)~\cite{Raffel2019-dq} for relevance-based re-ranking. These advanced methods, as utilized by Sarrouti et al.~\cite{Sarrouti2021-cq}, helped improve the accuracy of identifying pertinent articles. Evidence can also be sourced from citations in fact-checking websites and news articles, as demonstrated by Kotonya et al.~\cite{Kotonya2020-zy}, who utilized references from trusted sources to ensure accurate claim verification. Additionally, evidence retrieval can involve human input, where crowdworkers use tools like Google Search to gather evidence, focusing on credible sources such as government websites (e.g.,``.gov" or ``.mil" domains) or reputable medical sites~\cite{Mohr2022-bp}. To enhance relevance, queries were formulated from keywords extracted from the claim, specifically targeting authoritative databases like PubMed and other scholarly repositories. Retrieval accuracy was assessed using metrics such as precision (P@10), recall (R@10), and normalized discounted cumulative gain (NDCG@10)~\cite{Sarrouti2021-cq}, while inter-annotator agreement (IAA), such as Cohen's kappa, ensures consistency in manual retrieval tasks~\cite{Wuehrl2024-ro}.

Evidence matching is the process of identifying the sentences within retrieved documents that most effectively support or refute a claim, and subsequently forming the final evidence-sentence pairs. This task often employs semantic similarity techniques, such as cosine similarity between sentence embeddings, with S-BERT embeddings or SciSpacy embeddings tailored for biomedical texts~\cite{Deka2022-vj,Tan2023-oi,Kotonya2020-zy}. Wührl et al.~\cite{Wuhrl2022-em} utilized entity-relation triples to calculate semantic similarity, proposing methods that condensed claims based on these triples or the shortest sequence containing relevant entities. Natural language inference (NLI), a technique for determining whether one text logically supports, contradicts, or is neutral towards another, is also widely used for evidence matching. Wadden et al.~\cite{Wadden2020-kg} applied BERT-based models like RoBERTa-large, SciBERT~\cite{Beltagy2019-tw}, and BioMedRoBERTa to identify rationale sentences directly linked to the claim, while Liu et al.~\cite{Liu2020-bc} integrated SciBERT with a kernel graph attention network (KGAT)~\cite{Wang2019-hp} for fine-grained reasoning. Manual extraction methods also play a significant role in evidence matching. Sarrouti et al.~\cite{Sarrouti2021-cq} manually extracted relevant statements and Akhtar et al.~\cite{Akhtar2022-zj} utilized crowdsourcing to verify table relevance. In these studies, evaluation metrics for automated matching typically included precision, recall, and F1, while IAA measures, such as Krippendorff’s alpha, Fleiss’ kappa, and Randolph’s kappa, assess consistency in manual evidence matching~\cite{Akhtar2022-zj}.\\

\textbf{Building Claim Verification Systems}

Veracity prediction is the classification task of assigning an alignment label to each claim-evidence pair, typically categorizing claims as 3 labels: SUPPORTS, REFUTES, or NOT ENOUGH INFORMATION (NEI). Some datasets incorporate additional labels, such as the PUBHEALTH dataset, which uses four labels (TRUE, FALSE, MIXTURE, and UNPROVEN)~\cite{Kotonya2020-zy}, and the BEAR-FACT dataset, which includes five labels (SUPPORTED, PARTIALLY SUPPORTED, REFUTED, PARTIALLY REFUTED, and UNVERIFIABLE)~\cite{Wuehrl2024-ro}. BERT-based models, such as BioBERT~\cite{Lee2020-xg}, SciBERT~\cite{Beltagy2019-tw}, RoBERTa~\cite{Liu2019-yy}, and KGAT~\cite{Wang2019-hp}, were widely used across studies for this classification task. Classification performance was commonly evaluated with metrics like accuracy, precision, recall, and F1 score. Liu et al.~\cite{Liu2020-bc} trained the SciKGAT model on the SCIFACT dataset, which contains 1,409 claims with three labels (SUPPORTS, REFUTES, or NOINFO), achieving an F1 score of 0.5833 at the abstract level and 0.5048 at the sentence level. Similarly, Kotonya et al.~\cite{Kotonya2020-zy} fine-tuned SciBERT and BioBERT on PUBHEALTH, reporting F1 scores of 0.7052 and 0.6748, respectively. Wührl et al.~\cite{Wuehrl2024-ro} fine-tuned a RoBERTa model on BEAR-FACT, a dataset derived from Twitter posts with annotated fact-checking verdicts and structured subject-relation-object triplets. The model achieved an F1 score of 0.82 for the verifiable class but struggled with the unverifiable class, achieving only 0.27 F1. 

Validation and interpretation is the final step in fact-checking and claim verification, focusing on model explainability. This stage aims to clarify the reasoning behind predictions and increase trust and transparency in the model's conclusions. However, this step is often overlooked, with only a few studies incorporating explanation generation and human evaluation. Kotonya et al.~\cite{Kotonya2020-zy} proposed a hybrid explanation generation method using extractive-abstractive summarization, where evidence retrieved by their fact-checking pipeline was summarized into natural language justifications via a BERT-based model. Human annotators then assessed the coherence and relevance of these explanations to ensure they were logically consistent with the claim and its veracity label. In Tan et al.~\cite{Tan2023-oi}, human annotators evaluated claims generated from multiple-choice questions for qualities like fluency, contextuality, and faithfulness. While some studies included expert evaluations for claim quality or evidence matching, few employed specific metrics to measure interpretability. In future studies, effective validation and interpretation methods will be essential for creating more transparent fact-checking systems that can explain decisions comprehensively and reliably.

\subsection{Misinformation correction}

Misinformation correction has received limited attention in NLP due to the public and dynamic nature of misinformation. While error correction typically addresses errors in controlled systems with direct implications for patient outcomes, misinformation spreads rapidly across social media and public platforms, influencing large and diverse audiences. Correcting misinformation is further complicated by psychological barriers, such as motivated reasoning and confirmation bias, which lead individuals to reject corrections that conflict with their existing beliefs~\cite{Bode2018-nx}. Although recent advancements in error correction have leveraged LLMs to retrieve knowledge and correct errors in static datasets, misinformation correction demands a more dynamic approach. Countering false claims in real-time on public platforms requires generating responses that are not only timely and evidence-based but also tailored in tone and sensitivity to effectively engage the audience. This review highlights two studies that address these challenges by integrating evidence retrieval with response generation to produce polite, accurate, and audience-aware counter-misinformation responses.

He et al.~\cite{He2023-du} developed a reinforcement learning-based system called MisinfoCorrect specifically for misinformation correction, targeting COVID-19 vaccine misinformation. The study highlighted the critical role of generating polite, evidence-backed responses to effectively counter misinformation, addressing a key challenge in misinformation correction: user-generated responses were often uncivil or lacked substantiation, which could lead to arguments and erode trust. MisinfoCorrect aimed to overcome these barriers by producing respectful, friendly, and evidence-supported counter-responses that refuted false claims, improving the chances of reducing belief in misinformation. The system demonstrated how politeness not only fostered constructive discourse but also enhanced the credibility and impact of misinformation correction efforts. Evaluations of response quality focused on metrics such as politeness, refutation strength, evidence support, fluency, and relevance. Yue et al.~\cite{Yue2024-at} developed the Retrieval-Augmented Response Generation (RARG) framework for misinformation correction, focusing on generating evidence-backed counter-responses to COVID-19 misinformation. The system addressed common shortcomings in misinformation correction: a lack of supporting evidence and poor adaptability to domain shifts. RARG combined two key modules: evidence retrieval and evidence-based response generation. The evidence retrieval pipeline utilized a two-stage process (a coarse search with BM25 followed by fine-grained reranking using a dense retriever) to collect relevant evidence from over 1 million academic articles sourced from CORD-19 and LitCovid. The evidence-based response generation module employed reinforcement learning from human feedback (RLHF) to align LLMs for generating polite, factual, and relevant counter-responses. The evaluation metrics for this framework included refutation strength, factuality, politeness, claim relevance, and evidence relevance. By integrating retrieval and response generation, RARG consistently outperformed baseline models in both in-domain and cross-domain experiments. 

\subsection{Others}

Several studies explored other aspects of medical misinformation and provided complementary insights. Nabożny et al.~\cite{Nabozny2023-bg} proposed a semi-automatic strategy for identifying non-credible medical statements rather than directly classifying misinformation. This approach evaluated credibility based on factors like potential harm, misleading persuasion, and inconsistency with medical guidelines. Their definition of credibility had broader indicators including emotional language, unverifiable claims, and persuasion tactics, which were easier for machine learning models to detect. Models like LR and BioBERT demonstrated high precision, ranging from 0.835 to 0.986. Similarly, Zhou et al.~\cite{Zhou2020-mf} developed ReCOVery, a multimodal platform for assessing the credibility of COVID-19 news. This system integrated textual, visual, temporal, and network-based features from 2,029 news articles and 140,820 tweets. By employing the SAFE model~\cite{Zhou2020-vq}, a neural-network-based method proposed in their previous study that combines textual and visual features of news and evaluates the relevance between text and images for fake news detection, they achieved F1 scores of 0.833 for reliable news and 0.672 for unreliable news. Hossain et al.~\cite{Hossain2020-bc} introduced a stance detection dataset, COVID-Lies, which includes 6,761 tweets annotated for alignment with 86 curated COVID-19 misconceptions. Their stance detection task focused on classifying tweets into three categories: Agree, Disagree, or No Stance with respect to a given misconception. Compared with fact-checking tasks, this article emphasizes alignment rather than veracity.

Chin et al.~\cite{Chin2020-pr} focused on the characterization of misinformation through psycholinguistic analysis, sentiment analysis, and semantic representations, with a particular emphasis on inaccurate information regarding the HPV vaccine. Their study targeted misconceptions such as erroneous causal claims and toxicity myths found in online articles, news sources, and blogs. They found that false texts tend to be more narrative and emotionally negative, suggesting that such content is easier for readers to process and more likely to provoke negative sentiment. Cheng et al.~\cite{Cheng2021-vm} analyzed misinformation network evolution and predicted influential misinformation nodes. They use BERT embeddings from tweets to capture essential semantic and syntactic elements. These embeddings are then used to train a deep neural network (DNN) to predict which misinformation posts will become influential, which facilitates real-time intervention. 

\section{Hallucination}
\label{sec:hallucination}

Fifteen articles about hallucination detection and mitigation were included in this study. Table \ref{tab:hallucination} provides an overview of these articles. Most articles were in the general medical domain, but some focused on specific topics, such as ophthalmology~\cite{Hua2023-od} and pharmacovigilance~\cite{Muneeswaran2023-dn}. The source document type for hallucination-related tasks mainly came from question-answer pairs, either from medical examinations or publicly available QA datasets~\cite{Pal2023-nm,Ji2023-qq,Zakka2024-bi,Pal2024-ci,Muneeswaran2023-dn}. In addition, scientific articles~\cite{Pal2023-nm,Tang2023-zk,Pal2024-ci}, radiology reports~\cite{Van-Veen2023-fx,Liu2024-mr}, and patient-doctor dialogues~\cite{Qin2024-ts} were often used in text summarization and text generation tasks. Unlike error and misinformation detection, there were few datasets and studies directly related to hallucination detection. Many studies detected and evaluated hallucination using publicly available QA datasets listed in Table \ref{tab:hallucination}. Experts were usually included to evaluate the factual accuracy of AI-generated text. 

\begin{table}[H]
\tiny
\centering
\caption{Overview of NLP research about hallucination (Task: 1 = hallucination detection, 2 = hallucination mitigation)}
\label{tab:hallucination}
\begin{tabularx}{\textwidth}{@{}XXp{4em}p{10em}p{10em}p{8em}p{10em}p{5em}@{}}
\toprule
Ref. & Topic & Task & Document type & Dataset & Method & Metrics & Factuality evaluation \\
\midrule
Pal et al.~\cite{Pal2023-nm}         & general medical                                    & 1                                                                         & multiple-choice questions, PubMed abstracts                                                     & Med-HALT                                                                            & Text-Davinci, GPT-3.5, LlaMa-2, MPT, Falcon                               & accuracy, pointwise score                                                                      & automatic, expert                                      \\
Van Veen et al.~\cite{Van-Veen2023-fx}  & 	general medical  & 	1  & 	LLM-generated summaries  & 	Open-i, MIMIC-CXR, MIMIC-III, MeQSum, ProbSum	& GPT-4, GPT-3.5, FLAN-T5, FLAN-UL2, Vicuna, Llama-2  & 	BLEU, ROUGE-L, BERTScore, MEDCON, correctness, completeness, conciseness  & 	automatic, expert \\
Vishwanath et al.~\cite{Vishwanath2024-ki}   & 	general medical   & 	1	   & LLM-generated summaries   & 	MIMIC-IV	   & GPT-4o, Llama-3, Hypercube	   & incorrectness, specific-to-general   & 	expert \\
Yim et al.~\cite{Yim2024-qw}         & general medical                                    & 1                                                         & multiple-choice questions, open-ended responses, binary statements                              & internal dataset                                                                    & GPT-4, GPT-3.5, Llama2, PALM, BioMedLM, Dragon                            & accuracy, consistency, HumAgree, recovery, explain, ROUGE, BERTScore, BLEURT                   & automatic, expert                                      \\
Liu et al.~\cite{Liu2024-mr}         & general medical                                    & 1                                                         & multiple-choice questions, clinical summarizations, radiology reports, patient-doctor dialogues & MedQA, MedMCQA, PubMedQA, MIMIC-CXR, MIMIC-III, BC5-disease, NCBI-Disease, DDI, GAD & 9 general LLMs and 7 medical LLMs                                         & accuracy, F1, ROUGE-L, BLEU-4, faithfulness, comprehensiveness, generalizability, robustness   & automatic, expert                                      \\
Yang et al.~\cite{Yang2024-yc}  & 	general medical  & 	1  & 	LLM-generated medical abstracts  & 	Medline, PubTator  & 	Scorpius, ChatGPT, BioBART	  & writing fluency, context coherence, scientific faithfulness	  & automatic, expert \\
Hua et al.~\cite{Hua2023-od}       & ophthalmic subspecialties                          & 1                                                         & chatbot-generated scientific abstracts and references                                           & internal dataset                                                                    & GPT-3.5, GPT-4                                                            & DISCERN criteria (truthfulness, helpfulness, and harmlessness), hallucination rate, fake score & automatic, expert                                      \\
Tang et al.~\cite{Tang2023-zk}       & six clinical domains                               & 1                                                         & Cochrane reviews, LLM-generated summaries                                                       & internal dataset                                                                    & GPT-3.5, GPT-4                                                            & ROUGE-L, METEOR, BLEU, coherence, factual consistency, comprehensiveness, harmfulness          & automatic, expert                                      \\
Ji et al.~\cite{Ji2023-qq}          & general medical                                    & 1, 2                                                       & question-answering pairs                                                                        & PubMedQA, MedQuAD, MEDIQA2019, LiveMedQA2017, MASH-QA                               & Vicuna, Alpaca-LoRA, ChatGPT, MedAlpaca, Robin-medical                    & F1, ROUGE-L, Med-NLI, CTRL-Eval, query consistency, tangentiality, fact consistency            & automatic, expert                                      \\
Zakka et al.~\cite{Zakka2024-bi} 	  & general medical  & 	1, 2	  & clinical questions	  & ClinicalQA  & 	Almanac, Bard, Bing, GPT-4	  & factuality, completeness, preference   & 	expert \\

Pal et al.~\cite{Pal2024-ci}         & general medical                                    & 1, 2                               & multiple-choice questions, PubMed articles                                                      & MultiMedQA, MedQA, MedMCQA, PubMedQA, MMLU, Med-HALT, VQA Benchmark                 & Gemini Pro, 7 open source LLMs, 3 closed source LLMs                      & accuracy, pointwise score                                                                      & automatic, expert                                      \\
Qin et al.~\cite{Qin2024-ts} & pharmacy operations & 1, 2                                     & doctor-patient dialogues      & MedDG, KaMed & MedPH, LSTM, BERT, GPT-2, VRBot, DFMED                       & precision, recall, F1, BLEU, ROUGE, $\Delta$\text{GE, success rate}                                                & automatic                       \\
Xu et al.~\cite{Xu2024-ju}          & general medical                                    & 1, 2                                                  & medical questions, explanations of medical conditions, counterfactual scenarios                 & MedCF, MedFE                                                                        & MedLaSA                                                                   & efficacy, generality, locality, fluency                                                        & automatic, expert                                      \\
Muneeswaran et al.~\cite{Muneeswaran2023-dn} & pharmacovigilance                                  & 2                                                                                 & question-answering pairs                                                                        & PubMedQA, AEQA                                                                      & RAG, gpt-3.5-turbo, LLaMa-2                                               & faithfulness, accuracy, grade scores (by Auto-Grader)                                          & automatic                                              \\
Wang et al.~\cite{Wang2024-cx}        & general medical                                    & 2                                                                             & medical knowledge bases and guidelines                                                          & cMedKnowQA                                                                          & Alpaca, Bloom, ChatGPT                                                    & accuracy, helpfulness, harmlessness                                                            & automatic, expert                                      \\

\bottomrule
\end{tabularx}
\end{table}

\subsection{Hallucination detection}

In medical NLP, hallucination detection is critical for ensuring model reliability, particularly in tasks like question answering, summarization, and other text generation tasks. Hallucination detection focuses more on human evaluation of model-generated content rather than inaccuracies in existing texts. While error or misinformation detection generally uses classification metrics such as accuracy, precision, and recall, hallucination detection adds criteria like factual accuracy, coherence, faithfulness and hallucination rates.

In question answering, hallucination detection focuses on assessing the accuracy and consistency of model-generated responses to medical queries. Pal et al.~\cite{Pal2023-nm} introduced Reasoning Hallucination Tests (RHTs) through their Med-HALT framework. This framework, built on a diverse dataset from international medical exams, used tests like the False Confidence Test (FCT), None of the Above (NOTA) Test, and Fake Questions Test (FQT). These tests evaluated whether models like GPT-3.5 and LLaMA-2 could give accurate, non-hallucinatory responses. Med-HALT’s automatic metrics (e.g., accuracy) were paired with human evaluations to judge models on both factual accuracy and logical coherence. Similarly, Yim et al.~\cite{Yim2024-qw} explored how medical LLMs respond to slight changes in question-wording, which could impact answers dramatically. They tested models such as GPT-4 on both multiple-choice and open-ended formats. Their findings show that consistency often aligns with accuracy; however, even small variations in wording can shift the model’s performance. 

In summarization, hallucination detection examines whether models can generate faithful, comprehensive summaries of medical texts. Tang et al.~\cite{Tang2023-zk} tested LLMs like GPT-3.5 and GPT-4 on medical evidence synthesis using Cochrane Review abstracts across six clinical fields. The models had to capture key findings without introducing errors. Evaluations used both automatic metrics (e.g., ROUGE, METEOR, BLEU) and human assessments (e.g., coherence, factual consistency, comprehensiveness, harmfulness). Results showed that LLMs often missed important clinical details and sometimes generated overly confident summaries. Van Veen et al.~\cite{Van-Veen2023-fx} explored how adapted LLMs can outperform human experts in summarizing clinical texts, including radiology reports and progress notes. Their approach combined in-context learning, fine-tuning, and prompts tailored to medical summarization tasks. The study involved evaluations by ten physicians who assessed the clinical accuracy, relevance, and usefulness of the generated summaries compared to expert-written summaries. The results showed that GPT-4's summaries were preferred over expert summaries in 36$\%$ of cases and considered non-inferior in 45$\%$ of cases, with fewer hallucinations. Vishwanath et al.~\cite{Vishwanath2024-ki} focused on detecting and categorizing hallucination in clinical summaries generated by LLMs such as GPT-4o and Llama-3. They introduced a framework to classify hallucination into three main categories: (1) medical event inconsistency, which includes five subtypes such as errors in patient information, patient history, symptoms/diagnosis/surgical procedures, medicine related instructions, and follow-up; (2) chronological inconsistency, referring to discrepancies in the timeline of medical events; and (3) incorrect reasoning, where the logic or explanation associated with correct information is flawed. The study tested two automated hallucination detection approaches: an extraction-based system (e.g., Hypercube) and an LLM-based system (e.g., GPT-4o). While both methods showed promise, they also had limitations, such as overestimation or false positives. 

In other text generation tasks, hallucination detection focuses on ensuring the reliability of AI-generated academic and clinical content. Hua et al. ~\cite{Hua2023-od} studied the accuracy of GPT-3.5 and GPT-4 when generating scientific abstracts and references in ophthalmology. Using modified DISCERN criteria (including helpfulness, truthfulness, and harmlessness), they evaluated quality and calculated hallucination rates by verifying AI-generated references. They found high hallucination rates in citations, suggesting the need for caution when using AI-generated academic content without verification. Liu et al.~\cite{Liu2024-mr} introduced BenchHealth, a benchmark for testing LLM hallucination rates in healthcare across reasoning, generation, and understanding tasks. BenchHealth combined common metrics (e.g., accuracy, ROUGE-L, BLEU) with more specific measures of faithfulness, comprehensiveness, and robustness. This benchmark was used to evaluate general-purpose models like GPT-4 and fine-tuned medical LLMs like MedAlpaca. Results indicated that medical models often provided more faithful responses, while general-purpose models like GPT-4 offered more detailed answers with a higher risk of hallucination. Yang et al.~\cite{Yang2024-yc} developed Scorpius, a conditional text generation system that generates plausible yet hallucinated biomedical abstracts to test the vulnerability of medical knowledge graphs (KGs). Scorpius is built on an instruction-tuned LLM and guided by a scoring mechanism that refines synthetic abstracts to maximize their impact on KG rankings.  By conditioning on specific drug-disease pairs, Scorpius generated abstracts that significantly elevated the relevance of target drugs - 71.3$\%$ of drugs improved their rankings from the top 1,000 to the top ten positions. The quality of the Scorpius-generated abstracts was evaluated using metrics like perplexity, which showed better fluency and scientific consistency compared to ChatGPT. This study highlighted the potential risks posed by undetected hallucinated medical knowledge.

\subsection{Hallucination mitigation}

Hallucination mitigation focuses on preventing the generation of inaccurate information by LLMs. Unlike human-generated errors or misinformation, hallucinations are inherently model-generated inaccuracies. As these inaccuracies are produced during the text generation process, the priority shifts from post-hoc correction to proactive mitigation strategies. This is because correcting hallucination after generation not only adds an additional layer of complexity but also risks undermining trust in AI systems. Current research on hallucination mitigation mainly focuses on QA tasks, likely due to the availability of abundant QA datasets with gold-standard answers, which offer reliable benchmarks for evaluating model accuracy.

One key strategy in hallucination mitigation is RAG, which combines retrieval of relevant knowledge with the model’s generative abilities to produce more grounded responses~\cite{Xiong2024-dp, Gilbert2024-uv, Sahoo2024-bj}. Wang et al.~\cite{Wang2024-cx} used RAG to reduce hallucination in medical QA by integrating structured Chinese medical knowledge bases. They employed the cMedKnowQA dataset, using models such as LLaMA to enhance reliability by retrieving accurate information before generating responses. Evaluations included both accuracy metrics and manual assessments for helpfulness and safety, showing improved response faithfulness. Similarly, Muneeswaran et al.~\cite{Muneeswaran2023-dn} applied a multi-stage RAG framework to support biomedical inquiries on PubMedQA and an internal dataset focused on drug safety. Their approach improved GPT-3.5-turbo’s faithfulness and accuracy by over 15$\%$, employing rationale generation and verification to increase transparency and user trust. Zakka et al.~\cite{Zakka2024-bi} developed Almanac, a retrieval-augmented clinical language model evaluated on the ClinicalQA dataset, consisting of 314 open-ended clinical questions. By integrating curated medical resources such as PubMed, UpToDate, and BMJ Best Practices, Almanac achieved 91$\%$ citation accuracy and outperformed baseline models (ChatGPT-4, Bing, and Bard) on metrics of factuality, completeness, and user preference. Its adversarial safety mechanisms, including scoring query-context matches, prevented harmful outputs and ensured robust grounding of responses. Beyond RAG, Ji et al.~\cite{Ji2023-qq} introduced a self-reflection approach where models analyzed and verified their initial outputs before finalizing responses. This method used a hybrid of medical sources for validation and included both automatic accuracy measures and human expert evaluations, resulting in a significant reduction of hallucination in their QA tasks. 

Another method, CoT prompting, guides models through logical reasoning steps, often combined with ensemble refinement, where multiple answers are generated and refined~\cite{Sahoo2024-bj}. Pal et al.~\cite{Pal2024-ci} used CoT and ensemble methods to evaluate Google’s Gemini model across medical reasoning benchmarks. Their results showed that stepwise reasoning and answer refinement reduced hallucination rates and enhanced the reliability of diagnostic recommendations. 

Finally, model editing allows for precise modifications to the model’s knowledge without retraining, thereby reducing hallucination for specific topics. Xu et al.~\cite{Xu2024-ju} applied model editing to improve factual accuracy in medical LLMs, using their MedLaSA framework to edit both factual and explanatory knowledge. They evaluated the edited models with newly developed benchmarks, measuring fluency, locality, and efficacy, and demonstrated that targeted editing substantially improved response accuracy while preserving unrelated knowledge.

An intriguing perspective is presented by Qin et al.~\cite{Qin2024-ts}, who explored ``patient hallucination" in doctor-patient dialogues. These hallucinations were defined as discrepancies between the symptoms expressed by the patient and their actual health conditions. They often arose from patients’ lack of medical knowledge, anxiety, or miscommunication, which potentially led to inaccurate or contradictory information during consultations. To tackle this issue, Qin et al. introduced MedPH, a medical dialogue generation framework that integrated both hallucination detection and mitigation. The detection module employed graph entropy analysis on a dynamic dialogue entity graph to identify three types of patient hallucination: isolated entities, denial of critical entities, and self-contradictions. For mitigation, MedPH generated clarifying questions informed by the hallucination-related context, guiding patients to articulate their conditions more accurately. Experimental results on medical dialogue datasets demonstrated that MedPH outperformed baseline models in both entity prediction and response generation tasks, significantly reducing hallucination rates while maintaining response quality. This approach highlights the importance of addressing patient-provided inaccuracies to enhance the reliability of medical dialogue systems.

\section{Discussion}
\label{sec:discussion}

Methods applied to errors, misinformation, and hallucination show both similarities and differences. For example, classification-based error and misinformation detection often rely on machine learning and neural network models. However, while misinformation detection typically relies on social media or health news data, error detection is more focused on clinical text and medical reports. Furthermore, error correction in clinical texts primarily deals with lexical or phrase-level changes, while misinformation tasks emphasize fact-checking alongside evidence retrieval. In the context of managing hallucination in generative models, strategies from error and misinformation tasks can be adapted. For instance, insights from error detection, such as categorizing error types and employing structured prompts, can help guide models toward relevant content and reduce generative errors. Additionally, retrieval-augmented approaches in error correction, where models retrieve information from verified medical sources, offer a way to effectively ground responses. Similarly, misinformation detection, with its emphasis on evidence retrieval and multi-step verification, provides techniques directly applicable to validating generative text and minimizing hallucination. Techniques like similarity checks and ensemble methods further enhance alignment with reliable sources. 

The ultimate goal of addressing each type of inaccuracy reflects its specific context and challenges. For errors, correction is the most critical, as errors directly affect patient safety and healthcare outcomes. The ability to not only detect but also rectify incorrect dosages, diagnoses, or procedural details ensures accurate clinical decision-making. For misinformation, detection is often sufficient because the primary objective is to identify and flag inaccurate or harmful content before it spreads widely and influences public health behaviors. Once misinformation is flagged, further intervention can often be left to human reviewers or public health entities. In contrast, hallucination mitigation is essential for generative models because hallucinations are introduced during the text generation process. Correcting hallucination post-generation is not only resource-intensive but also risks eroding trust in AI systems. Mitigation strategies aim to ensure that models generate accurate information from the outset, making them more reliable for high-stakes applications like healthcare.

In the following subsections, we will highlight the challenges, limitations, and future directions of the methods applied to errors, misinformation, and hallucination.

\subsection{Errors}
\textbf{Privacy concerns.} Since the data used for error detection and correction mainly comes from clinical notes, it is often subject to privacy regulations such as HIPAA. The limited availability of publicly accessible clinical datasets further restricts the generalizability and reproducibility of models, as most are trained on proprietary datasets that are not accessible to the wider research community.

\textbf{Interoperability challenges.} Interoperability challenges arise because healthcare systems differ significantly in terminology, format, and documentation standards. This lack of uniformity means that models trained on data from one system may perform poorly in another. Another challenge is inconsistency in documentation.  Variations in how information is recorded across different clinical settings can lead to discrepancies. For instance, annotation inconsistencies, such as those found in National Violent Death Reporting System (NVDRS) dataset, can lead to errors in model predictions due to non-standardized coding or data input by human annotators~\cite{Wang2024-wn}. Such variability makes it challenging for NLP models to identify error patterns accurately across different datasets. 

\textbf{Synthetic errors.} Many existing clinical NLP datasets introduce synthetic errors, which have several notable drawbacks. Synthetic errors may fail to capture the complexity of real-world errors, as these errors are often context dependent. Furthermore, synthetic errors can introduce changes that do not reflect genuine mistakes, as they lack verification from medical experts. For example, substituting terms related to diagnosis or treatment with similar-sounding terms may produce new phrases that are still clinically accurate, potentially misleading the model during training. Another significant challenge is the difficulty of finding experts to validate errors, as this process often depends on the specific medical specialty and can involve varying levels of interpretation and correctness. 

\textbf{Need for multimodal analysis.} Some text-based errors can only be identifiable when analyzed alongside other data types, such as imaging, lab results, or genomic data. For example, detecting a finding error in a radiology report may require cross-referencing with radiology images to confirm or refute the documented finding. 

\textbf{Future directions.} To address these limitations, future work should focus on building datasets that capture naturally occurring errors in clinical text without violating privacy regulations. Additionally, using multimodal models that integrate text with imaging or lab results could enhance error detection accuracy, especially for complex cases where single-modality analysis is insufficient.

\subsection{Misinformation}
\textbf{Definitions.} The distinction between unintentional misinformation and disinformation is often overlooked in existing NLP studies. Without a clear understanding of the intent behind the content, it becomes difficult to assess the potential harm of the content and fully grasp its impact on public health. In clinical practice, failing to recognize intent can misguide interventions. A patient who misinterprets vaccine guidance due to low health literacy might benefit from personalized education~\cite{michel2021education}, while combating anti-vaccine disinformation requires institutional efforts like public rebuttals or platform moderation~\cite{swinford2024covid}. At a societal level, disinformation can exacerbate health inequities by targeting vulnerable populations. For example, during the COVID-19 pandemic, disinformation campaigns disproportionately affected minority communities by spreading false claims about vaccine safety~\cite{hildreth2021targeting}. Without intent-aware models, NLP systems risk treating all inaccuracies as equal, which can lead to blunt mitigation strategies that fail to address the root cause of the misinformation.

\textbf{Data sources and modalities.} Most datasets focus primarily on news articles and social media, which limits the scope of misinformation detection. Other sources, like podcasts, health advertisements, and patient information brochures remain underrepresented. Misinformation often involves multiple modalities, with images, videos, and graphics enhancing its impact. Addressing this requires expanding datasets to include these additional sources and modalities. 

\textbf{Challenges in fact-checking.} Although tools for automatic evidence retrieval and matching exist, finding high-quality evidence and accurately verifying claims still require significant input from experts. This reliance on expert labor limits the scalability of fact-checking systems, as it becomes difficult to handle large volumes of claims efficiently. Models that rely solely on claims as input, rather than claim-evidence pairs, tend to perform poorly. Furthermore, claims are typically brief statements that lack the full context provided by the original source. For example, the claim ``Regular exercise reduces the risk of chronic diseases" cannot be accurately verified without additional contextual information, such as the specific types of exercise, the chronic diseases being referred to, or details about the study, population group, or timeframe that support the assertion. This absence of contextual information limits the model’s ability to accurately assess the factuality of claims, as many claims require their original context to determine accuracy. Additionally, the factuality of some claims changes over time. For instance, the claim ``Vaccines for COVID-19 are not available to the public" was accurate in early 2020, but by late 2020, vaccines were authorized for emergency use, and by 2021, they were widely available. Using outdated evidence to verify such claims would lead to incorrect assessments, especially in rapidly evolving fields like the pandemic.

\textbf{Granularity of labels.} Misinformation exists at various levels of granularity. Some forms are not entirely false but are misleading or partially correct, making them difficult to identify and categorize using binary labels like ``real" or ``fake". For instance, a claim like ``Vitamin C cures the common cold" contains a kernel of truth, as vitamin C can support immune health~\cite{Carr2017-rl}, but it exaggerates the evidence, leading readers to believe it is a definitive cure. This highlights the need for more nuanced labeling systems that consider the degree of correctness and the potential harm such information could cause.

\textbf{Future directions.} Future research should focus on developing comprehensive, multimodal datasets that include diverse sources and formats. Improving context independence of claims in fact-checking tasks will enhance adaptability. NLP models should prioritize understanding the varying impact of misinformation on different subpopulations, focusing on vulnerable groups. To better differentiate unintentional misinformation from disinformation, future efforts should also prioritize the creation of benchmarking datasets annotated for intent (e.g., commercial, political, or sensational motives), which can guide the development of intent-aware classification models. Evaluating and implementing effective strategies and policies to prevent and mitigate health misinformation will advance both model performance and public health outcomes~\cite{Office-of-the-Surgeon-General-OSG-2021-xl}.

\subsection{Hallucination}
\textbf{Nature of LLMs.} Hallucinations in generative LLMs also have notable challenges. First, the inherent nature of these models makes it difficult to fully prevent outputs that sound coherent but lack factual accuracy~\cite{Guerreiro2023-re,Ji2022-wm,Gunjal2023-xk}. 

\textbf{Dataset scarcity.} The scarcity of standardized medical datasets for hallucination detection limits model development and benchmarking. There is a need for diverse, multinational datasets that cover various medical contexts and testing modalities~\cite{Pal2023-nm}. 

\textbf{Evaluation challenges.} Evaluation of hallucination often relies heavily on expert judgment, which introduces several challenges. Experts are tasked with assessing the factuality, coherence, and relevance of model-generated outputs, but these evaluations can be subjective and vary depending on the individual’s expertise and interpretation of the content. This subjectivity results in inconsistent evaluations across studies. Moreover, the cost of engaging medical experts, who are already in high demand, makes large-scale evaluations resource-intensive and impractical for many projects~\cite{Wang2023-mp}. The lack of unified evaluation guidelines further exacerbates these issues. Current metrics used in hallucination evaluation vary widely, ranging from automatic measures like BLEU and ROUGE to human assessments of factual accuracy, coherence, and comprehensiveness~\cite{Tam2024-cq}. However, these metrics are not always aligned. The absence of standardized, objective, and scalable evaluation frameworks presents a significant barrier to the development and validation of hallucination detection methods.

\textbf{Model challenges.} The effectiveness of hallucination detection also depends on the specific models being used. The rapidly evolving model landscape introduces additional difficulties, as new architectures and training techniques may require continual adaptation of detection methods and evaluation standards. This inconsistency hampers cross-study comparisons and reliable benchmarking. Building trust in LLMs for medical use requires greater transparency and explainability, which helps users understand the basis of model outputs. 

\textbf{Future directions.} In future studies, it is essential to develop datasets that capture real-world hallucination in healthcare settings. Advanced methods, such as using semantic entropy to detect confabulations, could improve detection by analyzing the stability of meaning across outputs~\cite{Farquhar2024-ed}. Plus, the variation in metrics across studies highlights the need for developing standardized guidelines and policies for consistent and reliable evaluation of hallucination in medical LLMs.

\section{Limitations of Scoping Review}
\label{sec:limitation}
While we conducted a comprehensive analysis of NLP techniques addressing medical errors, misinformation, and hallucination, some limitations of this review should be noted. First, although GPT-4o was used to assist with title and abstract screening and helped correct occasional human errors, it also introduced many false positives, recommending the inclusion of articles that did not meet the criteria upon further review. This highlights the need for careful human oversight when using LLMs as proxy reviewers. Second, despite efforts to include a broad range of studies, we might have missed relevant articles, particularly on COVID-19 misinformation, where we included only a selection due to the overwhelming volume of similar publications. Third, our focus on the medical domain meant that recent technical advances in the general domain were not included. Leveraging these rapid developments in LLMs may significantly enhance NLP approaches in various medical tasks. Fourth, while text-based NLP techniques were the primary focus, we did limited exploration on other data modalities, such as images, lab results, or audio. Multimodal studies are becoming increasingly important as the availability of diverse health data grows, and integrating text with other modalities has the potential to significantly enhance the detection and correction of medically inaccurate information. Lastly, the review lacks a consistent framework for evaluating each type of inaccurate information, as studies varied widely in their methods and metrics. Establishing standardized evaluation protocols would facilitate cross-study comparisons and improve the reliability of NLP models in medical applications.

\section{Conclusion}
\label{sec:conclusion}

This review underscores the progress made in using NLP to tackle medically inaccurate information, including errors, misinformation, and hallucination. With advancements in machine learning and LLMs, NLP has proven to be a valuable tool for tasks such as error detection and correction, misinformation detection, fact-checking, and addressing hallucination. However, several challenges remain, particularly regarding data privacy, synthetic data, contextual understanding, granularity levels, and standardization of evaluation metrics.

Integrating NLP into healthcare systems demonstrates potential to improve patient safety and public trust in medical information. Yet, ensuring the reliability and transparency of these technologies requires further research. Key priorities include developing multimodal datasets that better represent real-world complexities, improving methods to account for context, and establishing standardized evaluation frameworks for more consistent assessments of model performance. Collaboration among technologists, healthcare providers, and policymakers is essential to ensure the deployment of ethical, accurate, and robust NLP solutions.

Looking ahead, the field should explore the dual role of LLMs in mitigating and generating inaccuracies, particularly through more effective hallucination detection and mitigation strategies. By aligning technical innovations with healthcare needs, future research can advance the accuracy and reliability of medical information systems, contributing to enhanced patient outcomes and public health communication.

\section*{Acknowledgments}
This work was supported by the National Institutes of Health (NIH)—National Cancer Institute (Grant Nos. 1R01CA248422-01A1), National Library of Medicine (Grant No. 2R15LM013209-02A1), and National Center for Advancing Translational Sciences (Grant No. UL1 TR002319). The content is solely the responsibility of the authors and does not necessarily represent the official views of the NIH.

\newpage

\section*{Supplementary Information}
\label{sec:supplementary information}

\renewcommand{\thetable}{S\arabic{table}}
\setcounter{table}{0}

\begin{table}[H]
\tiny
\centering
\caption{Overview of the search queries and notes for each database}
\label{tab:search_query}
\begin{tabularx}{\textwidth}{@{}lXp{16em}@{}}
\toprule
Database & Search query & Note\\
\midrule
PubMed              & (("error"[Title/Abstract] OR "misinformation"[Title/Abstract] OR "disinformation"[Title/Abstract] OR "hallucination"[Title/Abstract] OR "misleading information"[Title/Abstract]) AND ("medical"[Title/Abstract] OR "health"[Title/Abstract] OR "healthcare"[Title/Abstract] OR "clinical"[Title/Abstract] OR "medicine"[Title/Abstract] OR "medication"[Title/Abstract]) AND ("natural language processing"[Title/Abstract] OR "NLP"[Title/Abstract] OR "text mining"[Title/Abstract] OR "LLM"[Title/Abstract] OR "large language models"[Title/Abstract] OR "chatbots"[Title/Abstract])) & 488 papers were retrieved based on the search query                                                                 \\\\
IEEE Xplore         & ((("error" OR "misinformation" OR "disinformation" OR "hallucination" OR "misleading information") AND ("medical" OR "health" OR "healthcare" OR "clinical" OR "medicine" OR "medication") AND ("natural language processing" OR "NLP" OR "text mining" OR "LLM" OR "large language models" OR "chatbots")))                                                                                                                                                                                                                                                                               & 449 papers were retrieved based on the search query                                                                 \\\\
ACM Digital Library & [[Abstract: error] OR [Abstract: misinformation] OR [Abstract: disinformation] OR [Abstract: hallucination] OR [Abstract: "misleading information"]] AND [[Abstract: medical] OR [Abstract: health] OR [Abstract: healthcare] OR [Abstract: clinical] OR [Abstract: medicine] OR [Abstract: medication]] AND [[Abstract: "natural language processing"] OR [Abstract: nlp] OR [Abstract: "text mining"] OR [Abstract: llm] OR [Abstract: "large language models"] OR [Abstract: chatbots]] AND [E-Publication Date: (01/01/2020 TO 11/30/2024)]                                       & 141 papers were retrieved based on the search query                                                                  \\\\
ACL Anthology       & ((error OR misinformation OR disinformation OR hallucination OR "misleading information") AND (medical OR health OR healthcare OR clinical OR medicine OR medication) AND ("natural language processing" OR NLP OR "text mining" OR LLM OR "large language models" OR chatbots))                                                                                                                                                                                                                                                                                                           & Only the most relevant 200 papers were identified for screening                                                     \\\\
Google Scholar      & ((error OR misinformation OR disinformation OR hallucination OR "misleading information") AND (medical OR health OR healthcare OR clinical OR medicine OR medication) AND ("natural language processing" OR NLP OR "text mining" OR LLM OR "large language models" OR chatbots))                                                                                                                                                                                                                                                                                                           & Only the most relevant 200 papers were identified for screening; 65 papers were manually added based on seed search 
\\
\bottomrule
\end{tabularx}
\end{table}

\begin{table}[H]
\tiny
\centering
\caption{Comparison of different text sources of medical error}
\label{tab:error_s1}
\begin{tabularx}{\textwidth}{@{}lp{4em}p{8em}Xp{16em}@{}}
\toprule
Ref. & Category & Source & Text & Note\\
\midrule
Eskildsen et al.~\cite{Eskildsen2020-yj} & 1 & Individual case safety reports  & It was reported that the patient had been using Levemir® PenFill® and NovoRapid® PenFill® for the last 2 years. It was reported that\textbf{ Levemir® and NovoRapid® were intentionally mixed in one syringe} (to minimize injections in pediatric patients).                                                                                                                                                                                                                                                   & Using a syringe to extract insulin from prefilled pens violates regulations in insulin administration, as it bypasses safety features designed to prevent dosing errors and contamination. \\\\
Wong et al.~\cite{Wong2020-fm}       & 1           & Incident reports                & Patient C was admitted to the emergency medicine ward (EMW) at 11:00 p.m. One day after Patient C’s admission (around 4 p.m.), a nurse found that another patient’s electronic patient record (ePR) was attached to Patient C’s medical file and discovered that\textbf{ the prescribed drugs shown on patient Ms. C’s medication administration record (MAR) did not belong to the patient’s usual medication list. However, the medication had already been administered as scheduled, according to the MAR.} & Incident reports document various types of errors, such as “wrong patient” and “wrong drug” labels in this case.                                                                           \\\\
Lee et al.~\cite{Lee2020-xg}       & 2           & Surgical pathologic records  & 
\makecell[tl]{\textbf{lug}, (right middle lobe), \textbf{wedgoe} resection:\\~- focal intra\\~- alveolar hemorrhage~\\~- no tumor present.} 
& This note contains spelling errors:``lug" should be ``lung", and ``wedgoe" should be ``wedge".                                                                                           \\\\
Abacha et al.~\cite{Abacha2024-se}     & 2          & Medical question-answering text & A 67-year-old man with type 2 diabetes mellitus and benign prostatic hyperplasia comes to the physician because of a 2-day history of sneezing and clear nasal discharge. He has had similar symptoms occasionally in the past. His current medications include metformin and tamsulosin. Examination of the nasal cavity shows red, swollen turbinates. \textbf{The patient is given diphenhydramine.}                                                                                                         & The correct medication is desloratadine. Diphenhydramine may not be appropriate due to its sedative effects, especially in older patients.                                                 \\\\
Pais et al.~\cite{Pais2024-di} & 2 & Prescriber directions & \makecell[tl]{Input direction: \\~- \textbf{1 po qhs}\\~- \textbf{500 mg priori to procedure}\\~- \textbf{tk 2–3 prn}\\~- \textbf{1 sprays intranasally 2 times per  day in each nostril}} & The desired output should be ``Take 1 capsule by mouth every night at bedtime", ``Take 1 tablet by mouth before procedure", ``Take 2 to 3 tablets by mouth as needed", and ``Instill 1 spray in each nostril twice daily", respectively \\
\bottomrule
\end{tabularx}
\end{table}

\begin{table}[H]
\tiny
\centering
\caption{Examples of claims and evidence in fact-checking datasets}
\label{tab:misinfo_s3}
\begin{tabularx}{\textwidth}{@{}lXXp{16em}Xp{20em}X@{}}
\toprule
Ref.            & Dataset   & Claim document type                                           & Claim text                                                                                                                                   & Evidence document type                      & Evidence text                                                                                                                                                                                                                                                                                                                               & Label            \\\\
\midrule
Wadden et al.~\cite{Wadden2020-kg}   & SCIFACT   & expert-written claims                                         & Cardiac injury is common in critical cases of COVID-19.                                                                                      & scientific abstracts                        & More severe COVID-19 infection is associated with higher mean troponin (SMD 0.53, 95\% CI 0.30 to 0.75, p  0.001)                                                                                                                                                                                                                           & SUPPORTS         \\\\
Sarrouti et al.~\cite{Sarrouti2021-cq} & HEALTHVER & claims returned by a search engine                            & COVID-19 is man-made in a lab.                                                                                                               & scientific articles                         & Recent research suggests that bats or pangolins might be the original hosts for the virus based on comparative studies using its genomic sequences.                                                                                                                                                                                         & REFUTES          \\\\
Mohr et al.~\cite{Mohr2022-bp}     & CoVERT    & Twitter/X (tweets)                                            & 5G networks cause covid.                                                                                                                     & scientific articles                         & There are two types of conspiracy associated with 5G-COVID-19. One version suggests that radiation from 5G lowers your immune system, which makes you more susceptible to the virus (Shultz, 2020). The idea that ...                                                                                                                       & REFUTES          \\\\
Kotonya et al.~\cite{Kotonya2020-zy}  & PubHealth & claims and cited sources from fact-checking and news websites & Under Obamacare, patients 76 and older must be admitted to the hospital by their primary care physicians in order to be covered by Medicare. & explanations by journalists                 & Obamacare does not require that patients 76 and older must be admitted to the hospital by their primary care physicians in order to be covered by Medicare.                                                                                                                                                                                 & FALSE            \\\\
Tan et al.~\cite{Tan2023-oi}      & Med-Fact  & multiple-choice question-answering datasets                   & Collagen fibers, elastic fibers, and reticular fibers comprise connective tissue.                                                            & multiple-choice question-answering datasets & ...Collagen fibers are interwoven with carbonhydrate-containing protein molecules called proteoglycans. Collectively, these materials are called the extracelluar matrix. Not only does the extracellular matrix hold the cells together to form a tissue, but it also allows the cells within the tissue to communicate with each other... & not-enough-info  \\
\bottomrule
\end{tabularx}
\end{table}

\newpage
\bibliographystyle{elsarticle-num}
\bibliography{ref.bib}

\end{document}